\documentclass[letterpaper, 10 pt, conference]{ieeeconf}  %

\IEEEoverridecommandlockouts                              %

\usepackage{caption}   
\usepackage{graphicx}  
\usepackage{caption}
\usepackage{subfigure} 

\usepackage{epsfig} %
\usepackage{mathptmx} %
\usepackage{times} %
\usepackage{amsmath} %
\usepackage{amssymb}  %
\usepackage{lineno}
\usepackage{cite}
\usepackage[breaklinks, colorlinks=true, linkcolor=blue, citecolor=blue, urlcolor=blue]{hyperref}

\usepackage[capitalize,nameinlink]{cleveref}
\crefname{section}{Sec.}{Secs.}
\Crefname{section}{Section}{Sections}
\Crefname{subsection}{Subsection}{Subsections}
\crefname{subsection}{Sec.}{Secs.}
\Crefname{table}{Table}{Tables}
\crefname{table}{Tab.}{Tabs.}
\Crefname{equation}{Equation}{Equations}
\crefname{equation}{Eq.}{Eqs.}
\usepackage{algorithm}
\usepackage{algorithmicx}
\usepackage{algpseudocode}  

\usepackage{pifont}
\usepackage[dvipsnames]{xcolor}
\usepackage{graphicx} %
\usepackage{amsmath}
\usepackage{amssymb}
\usepackage{adjustbox}
\usepackage{multirow}
\usepackage{colortbl}
\usepackage{cleveref}
\usepackage{url}
\usepackage{booktabs}

\newif\ifshowedits

\newcommand{\addeditor}[3]{%
  \definecolor{#1color}{rgb}{#3}
  \expandafter\newcommand\csname #1\endcsname[1]{%
  \ifshowedits
    {\color{#1color} ##1}%
  \else
    {##1}%
  \fi
  }%
  \expandafter\newcommand\csname #1rmk\endcsname[1]{%
  \ifshowedits
    {\color{#1color} {\bf [#2: ##1]}}
  \fi
  }%
  \expandafter\newcommand\csname #1rpl\endcsname[2]{%
  \ifshowedits
    {\color{#1color} ##1 \sout{##2}}
  \else
    {##1}
  \fi
  }%
}

\definecolor{MyGreen}{RGB}{0, 180, 0}
\definecolor{MyRed}{RGB}{180, 0, 0}
\definecolor{MyYellow}{RGB}{180, 180, 0}

\newcommand{\titletext}{{An End-to-End Framework for Optimizing Foot Trajectory and Force in Dry Adhesion Legged Wall-Climbing Robots}}

\showeditsfalse

\title{\LARGE \bf
\titletext
}

\author{Jichun Xiao$^{1}$ \quad Jiawei Nie$^{2}$ \quad Lina Hao$^{1}$ \quad Zhi Li$^{2}$
\thanks{$^{1}$ {The School of Mechanical Engineering and Automation, Northeastern University}, $^{2}$ {The State Key Laboratory of Synthetical Automation for Process Industries, Northeastern University}}%
}

\begin{document}

\maketitle
\thispagestyle{empty}
\pagestyle{empty}

\begin{abstract}
Foot trajectory planning for dry adhesion legged climbing robots presents challenges, as the phases of foot detachment, swing, and adhesion significantly influence the adhesion and detachment forces essential for stable climbing. To tackle this, an end-to-end foot trajectory and force optimization framework (FTFOF) is proposed, which optimizes foot adhesion and detachment forces through trajectory adjustments. This framework accepts general foot trajectory constraints and user-defined parameters as input, ultimately producing an optimal single foot trajectory. It integrates three-segment $C^2$ continuous Bezier curves, tailored to various foot structures, enabling the generation of effective climbing trajectories. A dilate-based GRU predictive model establishes the relationship between foot trajectories and the corresponding foot forces. Multi-objective optimization algorithms, combined with a redundancy hierarchical strategy, identify the most suitable foot trajectory for specific tasks, thereby ensuring optimal performance across detachment force, adhesion force and vibration amplitude. Experimental validation on the quadruped climbing robot MST-M3F showed that, compared to commonly used trajectories in existing legged climbing robots, the proposed framework achieved reductions in maximum detachment force by 28 \%, vibration amplitude by 82 \%, which ensures the stable climbing of dry adhesion legged climbing robots.
\end{abstract}

\section{Introduction}

A legged wall-climbing robot is a mobile robot with leg-like structures, designed to navigate vertical, inclined, or even inverted surfaces. These robots are used in tasks that pose risks or challenges for humans, such as inspection~\cite{li2022design}~\cite{murphy2011waalbot}, maintenance~\cite{gitardi2023trajectory}, cleaning~\cite{shi2022active}~\cite{tang2018switchable}, and search-and-rescue operations~\cite{zi2024intelligent}~\cite{nadan2024loris}. The adhesion methods for legged wall-climbing robots can be classified into magnetic adhesion~\cite{bandyopadhyay2018magneto}~\cite{hong2022agile}, electrostatic adhesion~\cite{sriratanasak2022tasering}~\cite{gu2018soft}, vacuum adhesion~\cite{gitardi2023trajectory}~\cite{hernando2022romerin}, and dry adhesion~\cite{kalouche2014inchworm}~\cite{li2022robust}. Compared to other adhesion types, legged robots utilizing dry adhesion offer distinct advantages. They achieve energy-efficient climbing by using materials that replicate biological adhesive mechanisms, enabling strong but easily reversible attachment without the need for complex mechanical systems. Dry adhesion is also characterized by quiet operation, minimal vibration, and leaving no residue, making it well-suited for use in sensitive environments such as laboratories, cleanrooms, and non-destructive testing scenarios. These advantages expand the robot's capabilities across various inspection and maintenance tasks.

As a special type of foot climbing robot, the trajectory of the foot end of the dry adhesion foot climbing robot consists of the support phase, the detachment phase, the swing phase and the adhesion phase trajectory. Among them, the support force of the support phase is completely provided by the adhesion force of the adhesion phase, so the design of the adhesion phase trajectory is the foundation for stable climbing. The task of the detachment phase is to remove the foot from the adherent surface and complete the forward movement in concert with the oscillating phase. However, the detachment process is often accompanied by large detachment forces. Therefore, the design of the detachment phase trajectory aims to minimize the detachment force, reduce dynamic interference during the detachment process, and ensure the stability of the overall climb.

Currently, most existing dry-adhesive legged climbing robots use conventional foot trajectory planning methods and rely on manual debugging experience to generate a segmented trajectory for a specific robot. These methods cannot optimize the large adhesion force and the small detachment force required for the stable movement of the foot because they do not take the adhesion force and detachment force of the foot as the optimization objectives, which affects the climbing performance and stability of the robot. Moreover, the constraints of different foot detachment structures are not considered, resulting in obvious limitations of the generated trajectories in terms of adaptability and scalability. In addition, the complex design of traditional methods makes it difficult to efficiently migrate and rapidly deploy them to different types of dry-adhesive legged climbing robots.

In order to address the limitations of traditional methods in terms of adaptability and scalability, an innovative Foot Trajectory and Force Optimization Framework (FTFOF) is proposed in this paper. The framework is based on a generalized acquisition platform and can generate optimal foot trajectories according to different task requirements by synergistically optimizing key performance indicators such as adhesion force, detachment force, and foot jitter at the foot. Compared with existing methods, FTFOF not only improves the adaptability and scalability of trajectories but also realizes rapid extension and portability to different types of dry-adherent legged climbing robots by designing user-friendly interfaces. This innovative framework provides core algorithmic support for dry adhesion legged climbing robots to achieve efficient and stable climbing in diverse motion scenarios. The main contributions of this paper are summarized as follows.

1) A foot optimization framework for dry adhesion legged climbing robots, termed the Foot Trajectory and Force Optimization Framework(FTFOF), is proposed. This framework adjusts the adhesion and detachment forces through foot trajectory modulation, aimed at obtaining a foot trajectory that is optimal for the given task.

2) A redundancy hierarchical strategy(RHS) has been proposed for finding the optimal solution within the Pareto front that meets the current task requirements by prioritizing multiple objectives and retaining the dominant solution using redundancy factors at each level.

3) A mobile generalized data acquisition platform has been constructed, which can acquire foot force data from different planes and different materials surfaces, and can form a corresponding sample set for different scenarios.

\section{Legged Climbing Robot Foot Optimization Framework}
\label{sec:Legged Climbing Robot Foot Optimization Framework}

\begin{figure*}[h]
    \centering
    \includegraphics[width=0.85\textwidth]{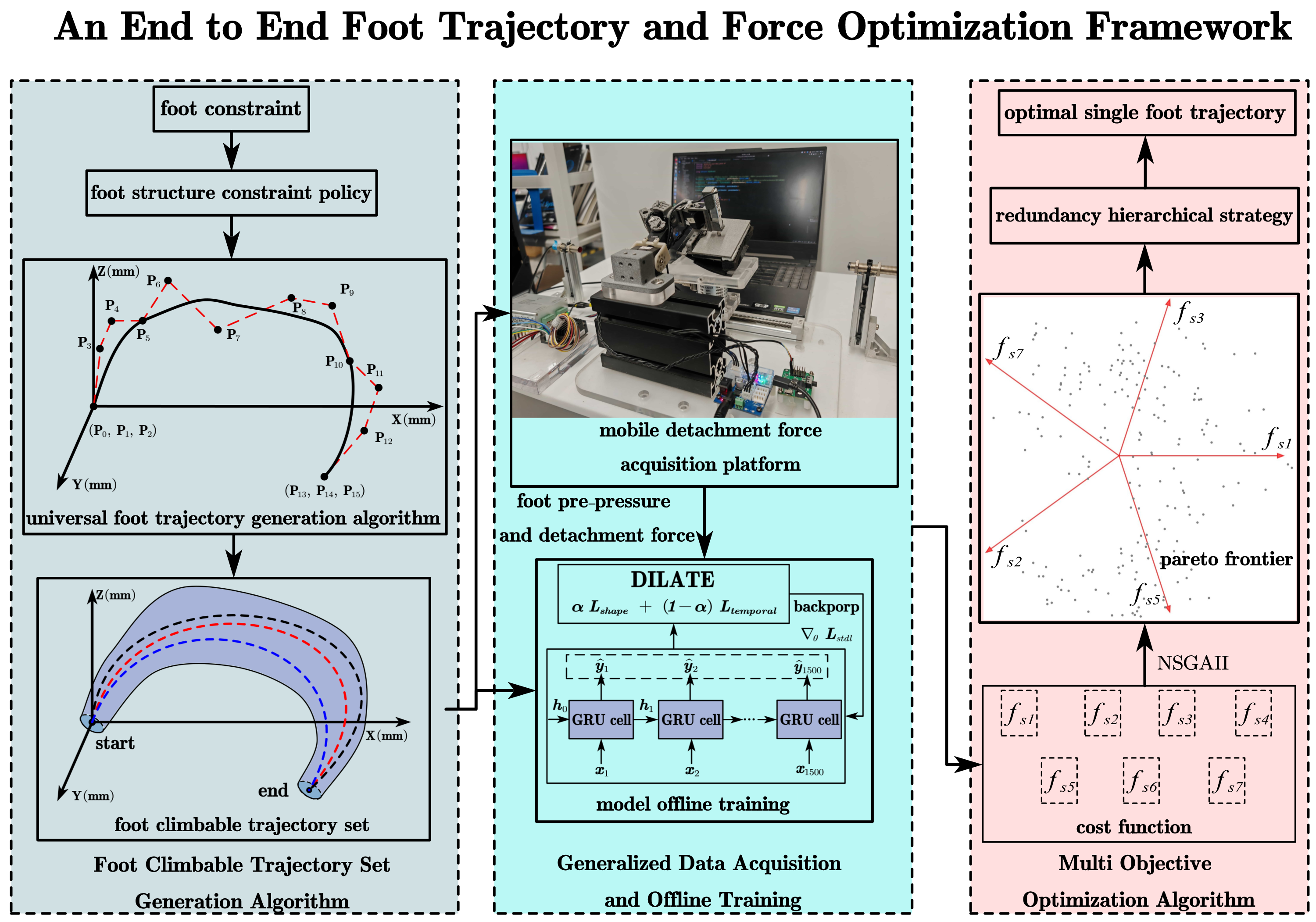}
    \caption{Legged Climbing Robot Foot Optimization Framework.}
    \label{fig:Legged_Climbing_Robot_Foot_optimization_framework}
\end{figure*}

This paper proposes an optimization framework for foot trajectories in dry adhesive legged climbing robots, termed the Foot Trajectory and Force Optimization Framework(FTFOF). The framework adjusts the foot adhesion and detachment forces through modulation of the foot trajectory to achieve optimal foot trajectories for the given task. The FTFOF proposed in this paper consists of three main modules: the foot climbable trajectory set generation algorithm module, the generalized data acquisition and offline training module, and the multi-objective optimization algorithm module, and the specific FTFOF algorithm framework is shown in Fig. \ref{fig:Legged_Climbing_Robot_Foot_optimization_framework}. Through the coupling between the three modules, the FTFOF algorithm is based on the multi-objective optimization algorithm, which can regulate the parameters of the foot adhesion force, foot detachment force, foot jitter and so on through the basic strategy function formulated in this paper, to obtain the optimal single foot trajectory that meets the current task requirements, which effectively improves the climbing stability of the dry-adhesive foot climbing robot.

\subsubsection{Foot Climbable Trajectory Set Generation Algorithm}
The Climbable Foot Trajectory Generation Module is designed to produce climbable foot trajectories suitable for various foot structures in dry adhesive legged climbing robots. The generation of these trajectories involves a two-step process. In the first step, a generalized foot trajectory for quadruped climbing robots is fitted using a three-segment $C^2$ continuous Bezier curve. In the second step, specific customized constraints are applied based on the unique foot structures of different climbing robots. The final output is a climbable foot trajectory that incorporates these specific customized constraints, including phases for detachment, adhesion, and swinging.

\subsubsection{Generalized Data Acquisition and Model Training}
In order to accurately construct an foot trajectory force model, the first task is to collect high-quality training and validation datasets. To this end, a mobile generalized data collection platform is specially developed in this paper to ensure that the collected foot force data can be applied to the needs of multiple tasks. The platform is capable of accurately acquiring foot force data on surfaces with different roughness and in different climbing scenarios (e.g., planar climbing and curved climbing data). In this way, training and validation sample sets corresponding to different tasks can be constructed to ensure the scalability of model training. In addition, the input of the model is the climbable foot trajectory data, while the output is the foot force trajectory data, which mainly includes the detachment force and pre-pressure data. This is because in the process of data acquisition, the actual collected foot force is the detachment force and pre-pressure data. The pre-pressure can fully characterize the adhesion force data while other constraints remain unchanged. And, ultimately, the detachment force and pre-pressure are also modeled separately, and the physical meanings of detachment force and adhesion force are different and cannot be directly integrated. Finally, in order to ensure the accuracy of the trained foot trajectory-detachment force mapping model and foot-end trajectory-pre-pressure mapping model, this paper adopts the GRU\cite{GRU} temporal memory network based on the DILATE\cite{DILATE} loss function for model training. The DILATE loss function can effectively deal with the shape and time distortion of the time-series data, while the GRU network can efficiently capture the dynamic features in the time-series data, thus ensuring the robustness and accuracy of the model in complex dynamic environments.

\subsubsection{Multi-Objective Optimization Algorithm}
The algorithm aims to obtain the optimal foot trajectory that best meets the task expectations. First of all, to ensure the stability of the robot during the climbing process and the specific task requirements, this paper designs a set of basic constraint strategy models for the climbing process of a foot-type climbing robot based on the foot trajectory-detachment force mapping model and the foot trajectory-pre-pressure mapping model. The constraint model comprehensively covers the key objective factors that affect the stable operation of the robot, laying a solid foundation for the optimization process. Secondly, for the specific task requirements, this paper adopts the classical NSGA-II\cite{NSGAII} multi-objective optimization algorithm. By setting multiple constraint strategies as objective functions and using Bessel control points as optimization variables, it can efficiently search and generate a Pareto front containing non-dominated solutions for all foot trajectories. This process not only ensures the diversity and comprehensiveness of the optimized solutions, but also provides rich candidates for subsequent trajectory selection. Finally, in order to quickly screen out the optimal trajectories of the foot that best meet the task requirements from the Pareto frontiers, this paper proposes a redundancy hierarchical strategy. This strategy can efficiently filter the optimal solution from many nondominated solutions based on the priority and specific requirements of the task. Through this strategy, it ensures that the robot can exhibit high adhesion, low detachment, and high stability when performing the task, thus realizing efficient and reliable climbing operations in complex environments.

The foot climbing trajectory set generation algorithm module lays the foundation for actual data acquisition. The general data acquisition and offline training module uses the foot climbing trajectory set to create the data set required for model training, which can realize the mapping models of various surface roughness and various climbing tasks. The multiobjective optimization algorithm takes the above model as the benchmark and obtains the required optimal trajectory, ensuring a stable climbing motion of the robot.

\section{Foot Climbable Trajectory Set Generation Algorithm}
\label{sec:Foot Climbable Trajectory Set Generation Algorithm}

The foot climbable trajectory set generation algorithm is designed to adapt to different foot structures of dry-adhesive foot climbing robots to generate foot trajectory sets that can satisfy climbing requirements. The algorithm consists of a universal foot trajectory generation algorithm and a foot structure constraint policy. The foot structure constraint strategy generates personalized foot trajectory constraints for different climbing robots with different foot structures, and combines with the generic foot trajectory generation algorithm to generate a set of climbable foot trajectories that can meet the task requirements.

\subsection{Universal Foot Trajectory Generation Algorithm}

The goal of the universal foot trajectory generation algorithm is to find the common feature applicable to the foot trajectories of all dry-adherent legged climbing robots, and use this common feature as the constraints of the Bezier curve to generate the formula of the generic foot trajectory algorithm. The common feature constraints of the foot trajectories can be summarized as the detachment decoupling constraint, the $C^2$ continuity constraint, and the start-end point smoothness constraint.

\subsubsection{detachment decoupling constraint}
\begin{figure}[h]   
    \centering
    \includegraphics[width=0.4\textwidth]{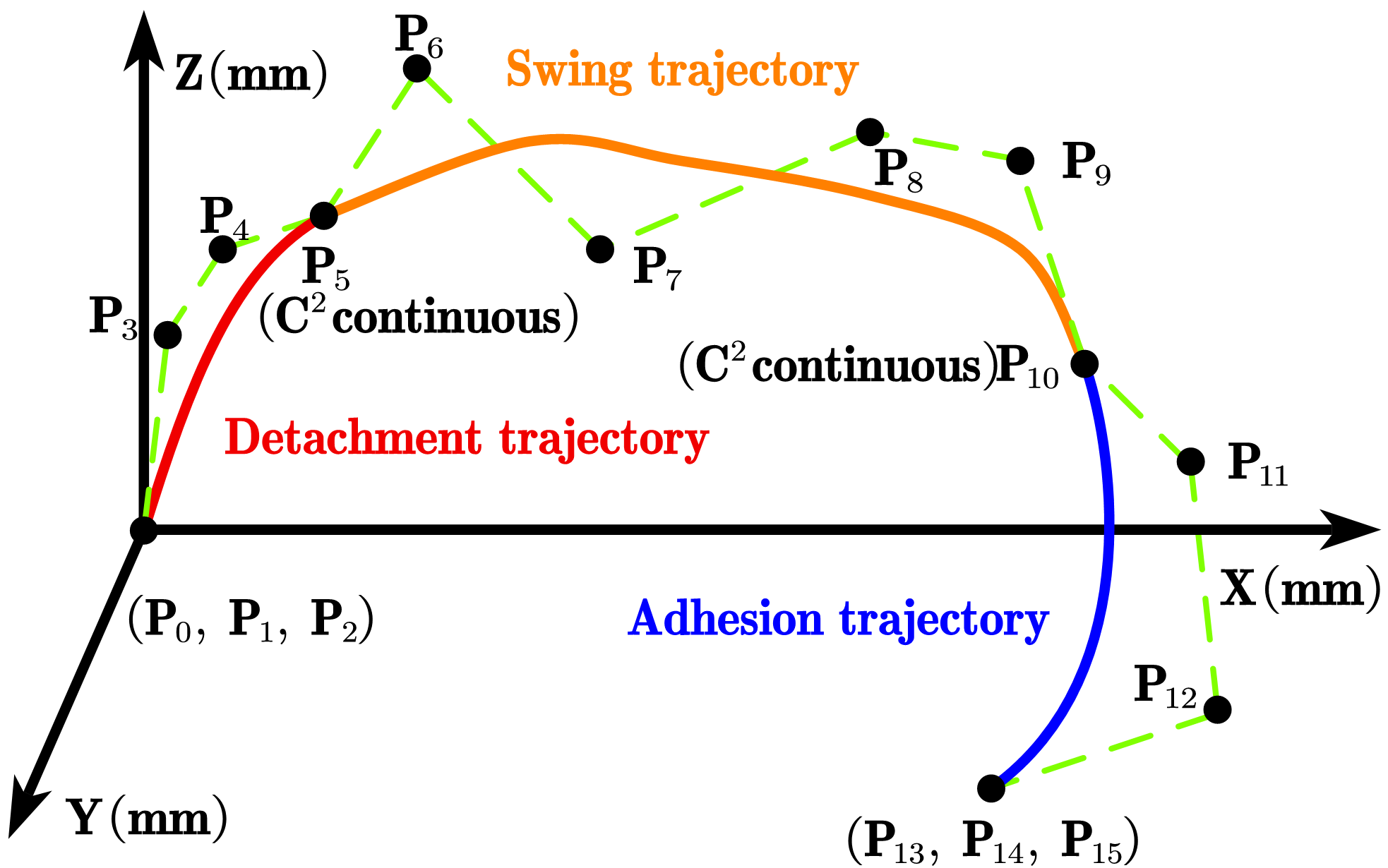}
    \caption{Foot Trajectories for Legged Climbing Robots.}
    \label{fig:bezier_curve_3D}
\end{figure}

In order to simultaneously exhibit detachment, oscillation and adhesion trajectories in the same foot trajectory, and to ensure that there is no interaction between detachment and adhesion, this paper proposes a strategy to decouple the detachment and adhesion trajectories. Specifically, three bezier curves are used to describe the detachment, swing and adhesion trajectories, respectively. Even if the subsequent c2 continuity constraints make the control points interact with each other, the mutual non-interference between the detachment and adhesion curves is guaranteed. The specific form of this decoupling strategy is shown in Fig. \ref{fig:bezier_curve_3D}. With this strategy, the regulation complexity problem brought by the high-order Bessel curves can be solved effectively.

\subsubsection{$C^2$ continuity constraint}
To ensure $C^2$ continuity at the splice of the three Bezier curves, it is required that the two neighboring trajectories are equal in position, velocity, and acceleration at the splice point. Therefore, during the splicing process, the six control points of the intermediate segment curves will be determined by the control points of the first and third segment curves, i.e., the Bezier trajectories that have been used at least $m \leq 5$ times. To satisfy the $C^2$ continuity of the overall trajectory and minimize the order of the Bezier curves, the three-segment fifth-order Bezier curves are chosen as the base curves in this paper. In this way, the detachment and adhesion curves can be kept independent of each other, and the order of the Bezier curves can be minimized to make the curves more controllable, as well as fitting the foot trajectories of all legged climbing robots. A generalized expression of a three-segment fifth-order Bezier position curve is as follows
\begin{equation} \label{eqa:c2_bezier_pos}
	\begin{cases}
		B_{0-5}(t_1) &= \displaystyle\sum\limits_{i=0}^{5} B_{i, 5}(t_1)P_{i} ,\ \ \ \ t_1 \in (0, 1] \\
		B_{5-10}(t_2) &=\displaystyle\sum\limits_{i=0}^{5} B_{i, 5}(t_2)P_{i+5},\ \ \ \ t_2 \in (0, 1] \\
		B_{10-15}(t_3) &=\displaystyle\sum\limits_{i=0}^{5} B_{i, 5}(t_3)P_{i+10} ,\ \ \ \ t_3 \in (0, 1]
	\end{cases}
\end{equation}
where $B_{0-5}, B_{5-10}, B_{10-15}$ denote the position curves of the first, second, and third Bezier curves, respectively, and the following analogies; $T$ is the total time required for the execution of the Bezier curves; and $t_1, t_2, t_3$ are the running times of each Bezier curve.

To realize the continuous and controllable operation of three-segment fifth-order Bezier curves at a given time, the running time of each Bezier curve is defined as
\begin{equation} \label{eqa:c2_bezier_traj_pos_time}
    \begin{cases}
        t_1 = \displaystyle\frac{t}{ T_1} , \ \ t \in (0, T_1] \\[3mm]
        t_2 = \displaystyle\frac{t-T_1}{T_2 - T_1} , \ \ t \in (T_1, T_2]\\[3mm]
        t_3 = \displaystyle\frac{t-T_2}{T_3 - T_2} , \ \ t \in (T_2, T_3]\\
    \end{cases}
\end{equation}
where $t$  is the real-time running time; $T_1, T_2, T_3$ are the end times when the execution of each Bezier curve is completed.

Further generalized expressions for three-segment fifth-order Bezier velocity and acceleration curve can be obtained
\begin{equation} \label{eqa:c2_bezier_speed}
	\begin{cases}
		\overset{(1)}{B_{0-5}}(t_1) = \displaystyle \frac{5!}{(5-1)!} \sum_{i=0}^{4} (\bigtriangleup ^{1}P_{i})B_{i,4}(t_1) \\[2.5mm]
		\overset{(1)}{B_{5-10}}(t_2) = \displaystyle \frac{5!}{(5-1)!} \sum_{i=0}^{4} (\bigtriangleup ^{1}P_{i+5})B_{i,4}(t_2) \\[2.5mm]
		\overset{(1)}{B_{10-15}}(t_3) = \displaystyle \frac{5!}{(5-1)!} \sum_{i=0}^{4} (\bigtriangleup ^{1}P_{i+10})B_{i,4}(t_3)  
	\end{cases}
\end{equation} 
and
\begin{equation} \label{eqa:c2_bezier_acc}
	\begin{cases}
		\overset{(2)}{B_{5-10}}(t_2) = \displaystyle \frac{5!}{(5-2)!} \sum_{i=0}^{3} (\bigtriangleup ^{2}P_{i+5})B_{i,3}(t_2) \\[2.5mm]
		\overset{(2)}{B_{10-15}}(t_3) = \displaystyle \frac{5!}{(5-2)!} \sum_{i=0}^{3} (\bigtriangleup ^{2}P_{i+10})B_{i,3}(t_3)
	\end{cases}
\end{equation}

According to Eqs. (\ref{eqa:c2_bezier_pos}), (\ref{eqa:c2_bezier_speed}), and (\ref{eqa:c2_bezier_acc}), combined with the condition of equality of position, velocity, and acceleration at the splicing of two adjacent trajectories, the corresponding $C^2$ continuity constraints can be obtained
\begin{equation} \label{eqa:c2_continuous_constraint_point}
    \begin{cases}
           P_{6} = 2P_{5} - P_{4} \\
           P_{9} = 2P_{10} - P_{11} \\
           P_{7} = 4(P_{5} - P_{4}) + P_{3} \\
           P_{8} = 4(P_{10} - P_{11}) + P_{12}
    \end{cases}
\end{equation}

\subsubsection{Start-End Point Smoothness Constraint}
To ensure the smooth running of the three-segment fifth-order Bezier curves at the starting point of the detachment trajectory and the end point of the adhesion trajectory, it is necessary to ensure that the velocities and accelerations of the Bezier curve at $t_1=0$ and $t_3=0$ are zero. According to Eqs. (\ref{eqa:c2_bezier_pos}), (\ref{eqa:c2_bezier_speed}), and (\ref{eqa:c2_bezier_acc}), combined with the condition of zero velocity and acceleration at the start and end point, the start-end point smoothness constraint can be obtained
\begin{equation} \label{eqa:lift-touch_constraint_point}
    \begin{cases}
        P_0 = P_1 = P_2 \\
        P_{15} = P_{14} = P_{13}
    \end{cases}
\end{equation}

Substituting the common features of the above foot trajectories as constraints into the Bezier curves, a generalized foot trajectory representation can finally be obtained.

\subsection{Foot Structure Constraint Policy}
The universal foot trajectory constraints described above have integrated disengagement, swing and adhesion phase trajectories and ensured $C^2$ continuity of the trajectories. However, given the different foot configurations of different dry-adherent foot climbing robots, this leads to different foot detachment methods and constraints. In order to provide a more versatile and portable framework for foot trajectory optimization, five user-specifiable forms of constraints are set in this paper. Users can input the constraint parameters according to the specific needs of the robot body and foot characteristics to ensure that the robot can successfully complete the detachment and adhesion process. The user-specified constraints include maximum foot position constraint, maximum foot velocity constraint, minimum detachment point constraint, transition control point constraint and trajectory shape constraint.

\subsubsection{Maximum Foot Position Constraints}
The maximum foot position needs to be restricted to avoid the foot's trajectory exceeding the robot foot's motion range. To ensure the algorithm's generality, the joint angle limitations, which are known to all dry adhesion foot climbing robots, are used as constraints. In this way, it can be applied regardless of the legged climbing robots' structure.

In this paper, a multi-objective optimization approach is used to indirectly solve for the foot's range of motion through the joint angle constraints. Specifically, the joint angle is used as an optimization variable, the joint angle limit is used as a boundary restriction of the optimization variable, the inverse kinematics of the robot leg is used as an objective function, and the multi-objective optimization algorithm is used to determine a suitable maximum foot position boundary $P_{pl}$. In addition, an inequality constraint on the x and z axis motions of the foot is required to ensure that it is desired that the foot can move far enough, rather than lift high enough, when solving for the foot position boundary. The specific optimization solution is as follows
\begin{equation} \label{eqa:foot_pos_limit}
    \begin{aligned}
		&\ \min P_{pl}(\theta) = - \mathrm{IK}(\theta) \\
		&\ \text { s.t. }  
			\begin{cases}
			& \displaystyle  \mathrm{IK}_{z}(\theta) - \mathrm{IK}_{x}(\theta) \le 0 \\
			& \min \theta_{i}  \leq \theta_{i} \leq \max \theta_{i} \\
		  	\end{cases}
	\end{aligned}
\end{equation}
where $IK(\theta)$ is the inverse kinematics model of the leg of a dry adherent legged climbing robot, $\theta_{i}$ is the joint angle, and $\min \theta_{i}$ and $\max \theta_{i}$ denote the limitation of the leg joint angle.

\subsubsection{Maximum Foot Velocity Constraints}
Similarly, for the robot maximum foot velocity constraint, it needs to be obtained by the above method to ensure the generality of the algorithm. However, the only difference is that the maximum foot velocity constraint should be input not only the joint angle constraint, but also the joint angular velocity constraint.

\begin{equation} \label{eqa:foot_vel_limit}
    \begin{aligned}
        &\ \min P_{vl}(\dot{\theta}) = \boldsymbol{J}(\theta, \dot{\theta}) \dot{\theta} \\
        &\ \text { s.t. }  
			\begin{cases}
			& \displaystyle \mathrm{IK}_{z}(\theta) - \mathrm{IK}_{x}(\theta) \le 0 \\
			& \min \theta_{i} \leq \theta_{i} \leq \max \theta_{i} \\
			& \min \dot{\theta_{i}} \leq \dot{\theta_{i}} \leq \max \dot{\theta_{i}} \\
		  	\end{cases}
    \end{aligned}
\end{equation}
where $\dot{\theta_{i}}$ is the angular velocity of the leg joints of the dry adhesion legged climbing robot, and $\min \dot{\theta_{i}}$ and $\max \dot{\theta_{i}}$ are the limits of the angular velocity of the leg joints and $J(\theta, \dot{\theta})$ is the Jacobi matrix for the legs.

\subsubsection{The Minimum Detachment Point Constraint}
In the design of the detachment trajectory, the initial stage of the detachment process is the most challenging part. To ensure a smooth initial detachment, this paper introduces the minimum detachment point $m_{d} = (h_{d}, l_{d}, w_{d})$ as a constraint for the initial detachment. The exact value of this minimum detachment point needs to be set according to the characteristics of the different foot structures.

\subsubsection{Transition Control Point Constraints}
The transition control points are the transition control point $P_{5}$ between the first detachment trajectory and the second swing trajectory, and the transition control point $P_{10}$ between the second swing trajectory and the third adhesion trajectory. $P_{5}$ is theoretically the end point of the detachment trajectory, and it must be larger than the minimum detachment point and smaller than the foot position boundary of the trajectory. $P_{10}$ is the end point of the swing trajectory and the limit end point to ensure complete detachment. Since the swing trajectory plays a demarcation role between the detachment and adhesion trajectories, when the detachment trajectory fails to achieve complete detachment, the remaining detachment process needs to be completed by the swing trajectory to ensure complete detachment. The generalized transition control point constraint equation is as follows
\begin{equation} \label{eqa:P5P10_limits}
	\begin{cases}
        m_{d} < P_{5} < P_{pl} \\
		c_{d} < P_{10} < P_{pl} 		
	\end{cases}
\end{equation}
where $c_{d}$ is the complete detachment point in the detachment trajectory.

\subsubsection{Trajectory Shape Constraints}
The trajectory shape constraints are additional constraints set in this paper, which are mainly set to prevent specific mechanisms from being included in the foot structure of a dry adherent legged climbing robot. It can be specified by the user on demand.

\section{Generalized Data Acquisition and Model Training}
\label{sec:Generalized Data Acquisition and Model Training}

The main task of generalized data acquisition and model training is to collect high-quality training and validation datasets, including data from different roughness surfaces and different climbing scenarios, using a mobile generalized data acquisition platform. The foot trajectory-detachment force model and foot trajectory-prepressure model trained by a GRU time-series network based on DILATE loss function were used.

\subsection{Mobile Generalized Data Acquisition Platform Design}
\begin{figure}[h]   
    \centering
    \includegraphics[width=0.45\textwidth]{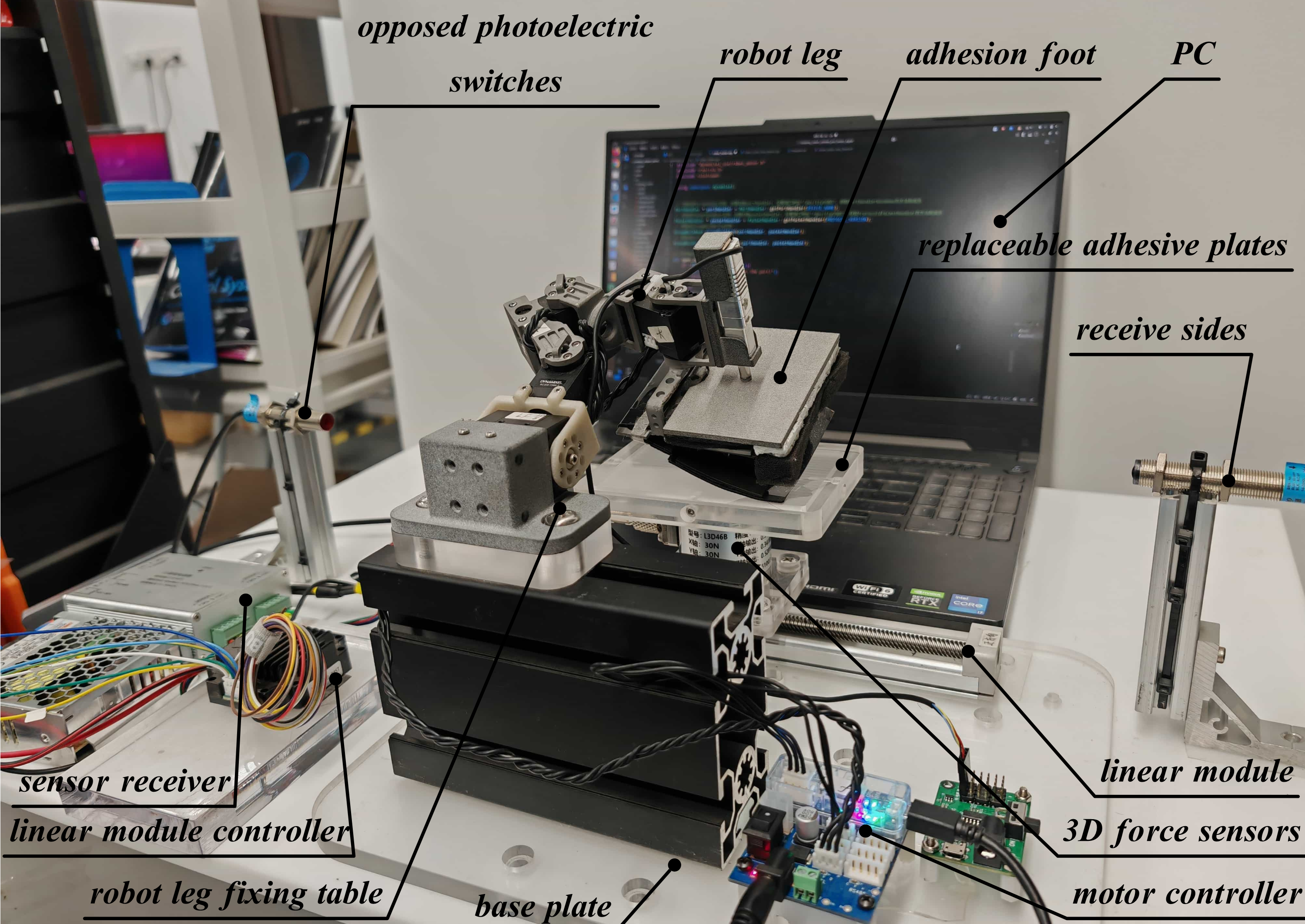}
    \caption{Mobile Detachment Force Acquisition Platform.}
    \label{fig:detachment_force_platform}
\end{figure}

In this paper, a mobile foot force generalized data acquisition platform is designed. The platform can make the force sensor move with the movement of the foot to ensure that the force point of the force sensor is always located in the center, to ensure the accuracy of the collected data. The structure of the mobile foot force eneralized data acquisition platform is shown in Fig. \ref{fig:detachment_force_platform}. The acquisition platform consists of a stationary base plate, a robot leg, a linear module, a linear module controller, 3D force sensors, a sensor receiver, replaceable adhesive plates, opposed photoelectric switches, a motor controller, and a control computer (PC). The fixed base plate is used to fix the robot leg fixing table and linear module, and is stably mounted on the table; the 3D force sensor is installed between the linear module and the replaceable adhering plate, and is used for real-time acquisition of the force data at the end of the foot; the PC is used for controlling the robot leg to complete the detachment movement, and real-time recording of the force data at the end of the foot sent by the sensor receiver; the transmitting and receiving sides of the opposed-type photoelectric switch are fixed on both sides of the linear module, and the mounting height is fixed to the linear module, with the transmitter and receiver sides fixed to the linear module. The transmitter and receiver of the butt-type photoelectric switches are fixed on both sides of the linear module, with the mounting height exceeding 2 mm above the top of the replaceable adhesion plate, and are used to receive the 0-1 signal of whether or not the adhesion foot has completed the detachment, and are used for controlling the linear module movement.

In addition, to enable the mobile foot force generalized data acquisition platform to adapt to more diverse climbing tasks and climbing modes, the adhesion plate of the platform is designed as a removable structure. Users can replace different types of adhesive plates according to the needs of specific tasks, to facilitate the collection of data applicable to different climbing tasks.

At the start of data acquisition, the robot leg was first secured to the leg fixation end and reset to the adhesion surface. Subsequently, an initial pre-pressure is applied to ensure that the foot adheres to the surface. Next, the foot is actuated to execute a climbable trajectory, and the foot force is recorded in real time by a 3D force sensor. Before the detachment movement, the transmitter and receiver of the photoelectric switch had zero photoelectric signals due to the blockage of the foot, and no command was issued to drive the linear module movement. However, when the foot is completely detached, the receiving end of the photoelectric switch receives the signal from the transmitting end, and then drives the linear module to move the force sensor to the final landing position of the foot, which ensures that the center of the foot and the center of the force sensor are also aligned with each other during the measurement process, and ensures the measurement accuracy.

\subsection{Model Training}

Traditional time series regression methods, such as linear autoregressive models (e.g., ARIMA models\cite{ARIMA} and exponential smoothing\cite{Exponential_Smoothing}), are usually used for the analysis of smooth time series. However, many time series in the real world present sudden changes in distribution, and the smoothness assumption is often not valid. Therefore, in this paper, we choose the GRU time-series prediction model with automatic feature extraction capability and multi-step prediction for predicting the data of foot trajectory and foot force.

\begin{figure}[h]   
    \centering
    \includegraphics[width=0.49\textwidth]{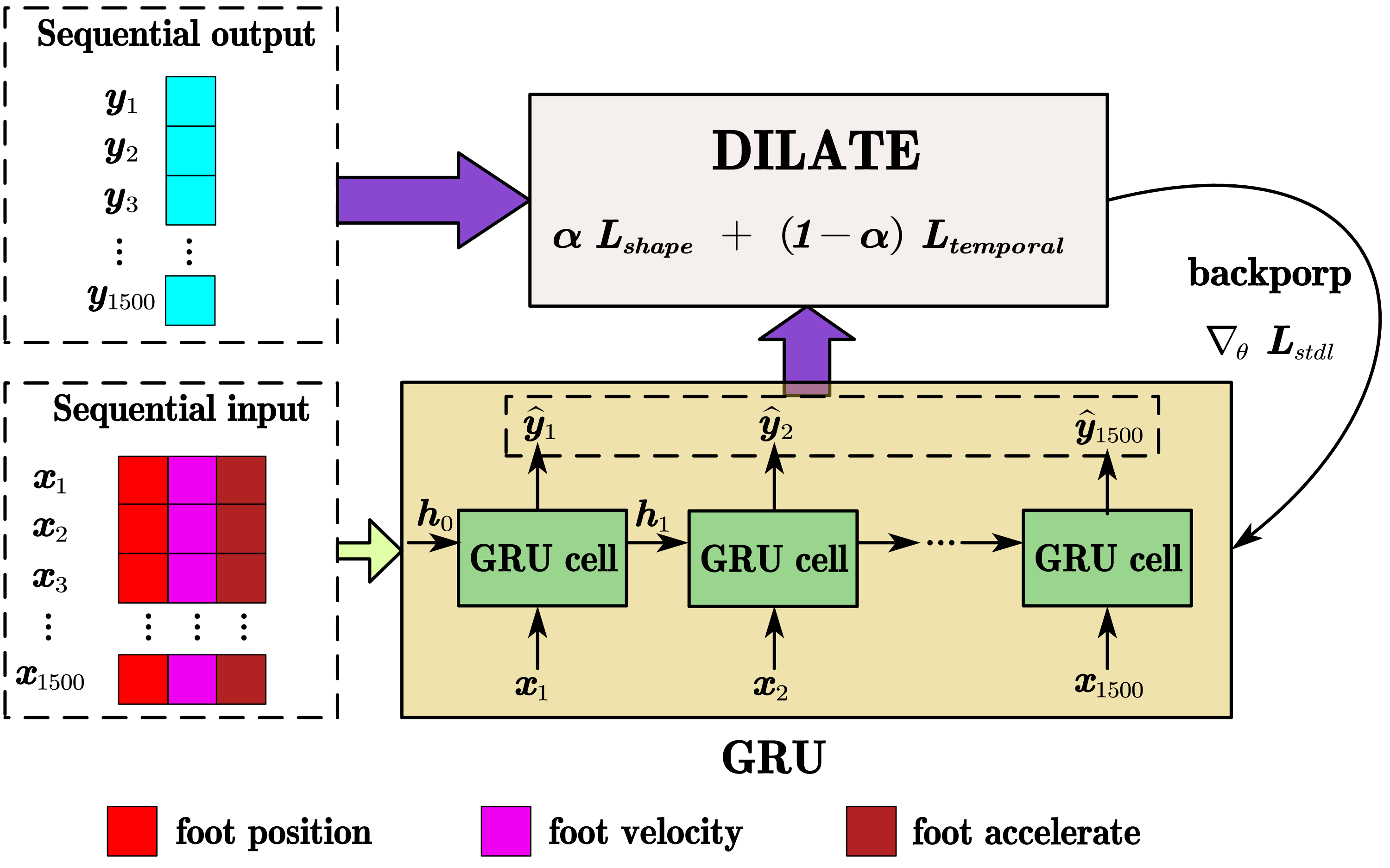}
    \caption{GRU Network Model Utilizing the DILATE Loss Function.}
    \label{fig:Deep_learning_framework}
\end{figure}

In addition, the real collected foot force data are usually disturbed and exhibit non-smooth time series characterized by abrupt changes in distribution, resulting in inconsistent pacing between the foot trajectory and force trajectory sequences. Currently, the commonly used loss functions, including MAE, MSE, and their variants, are difficult to deal with the above problems effectively. To solve this problem, DILATE\cite{DILATE} is chosen as the loss function. Through the calculation of shape loss and time distortion loss, not only can the similarity of trajectory shapes be obtained, but also the effect of time delay on the data can be taken into account to obtain a more accurate foot trajectory-detachment force model $F_{d}$ and foot trajectory-prepressure model $F_{p}$. The modeling framework is shown in Fig. \ref{fig:Deep_learning_framework}.

In model training, a sample set of foot climbable trajectories was used as input to the model, while the detachment force and pre-pressure trajectory data acquired via a mobile foot force acquisition platform were used as the model's output. 85\% of the dataset was used for training and 15\% as a validation set to ensure the model's generalization ability and accuracy across different datasets.
\section{Multi-Objective Optimization Algorithm}
\label{sec:Multi-Objective Optimization Algorithm}

To highlight the generality of the framework proposed in this study, seven fundamental constraint strategy models applicable to all dry adhesive legged climbing robots have been developed, based on deep learning-trained models of foot trajectory-adhesion force and foot trajectory-detachment force. Employing these strategy models as the cost function, with Bezier control points $P_{3x}$, $P_{4x}$, $P_{5x}$, $P_{10x}$, $P_{11x}$, $P_{12x}$, $P_{3z}$, $P_{4z}$, $P_{5z}$, $P_{10z}$, $P_{11z}$, and $P_{12z}$ as optimization variables, the paper seeks to determine the Pareto front that encompasses all non-dominated solutions. To identify the optimal trajectory for a specific task, we employ a redundancy-based stratification strategy to select the most appropriate trajectory from the Pareto front.

\subsection{Fundamental Strategies}

\subsubsection{Maximum detachment force constraint strategy}
This strategy is used to limit the maximum detachment force and is designed to minimize the transient effects on the climbing robot. The maximum detachment force is the maximum value that the robot can reach instantaneously and causes the robot to jitter, thus reducing climbing stability.

\begin{equation} \label{eqa:max_detachment_force}
	f_{s1} = \displaystyle \max(F_{d}(B_{p}))
\end{equation}
where $F_{d}$ denotes the foot trajectory-detachment force model derived from a deep learning algorithm. $ B_p$ represents the foot position trajectory, and $n$ is the number of foot position trajectory points.

\subsubsection{Average Detachment Force Constraint Strategy}
The strategy represents the average detachment force applied to the foot trajectory from the start of detachment to the complete detachment of the robot, and is intended to analyze the overall detachment effect. The specific constraints take the following form
\begin{equation} \label{eqa:sum_detachment_force}
	f_{s2} = \frac{\displaystyle \sum_{i=0}^{n}[F_{d}(B_p)]_{i}}{n}
\end{equation}

\subsubsection{Adhesion Performance Constraint Strategies}
This strategy aims to ensure precise control of the adhesion performance of the foot, especially the pre-pressure, which directly determines the amount of adhesion force. In addition, the strategy uses an absolute value to indicate whether the pre-pressure is too high or too low, which means that the adhesion performance is compromised. The specific constraints are in the following form
\begin{equation} \label{eqa:adhesion_force}
	f_{s3} = \displaystyle |\max(F_{p}(B_{p})) - a|
\end{equation}
where $F_{p}$ is the foot trajectory-prepressure model; a is the optimal pre-pressure value, which in this paper refers to the initial pre-pressure used during data acquisition, as the optimal pre-pressure value to ensure that the adhesion performance of the foot is in an optimal state.

\subsubsection{Foot Lift Height Constraint Strategy}
This strategy aims to limit the lifting height of the foot, to reduce the energy consumption of the robot during the climbing process, and to improve the climbing efficiency. Secondly, the strategy can also be used to constrain tasks that are limited by the lifting height of the foot. The specific constraints are as follows
\begin{equation} \label{eqa:traj_height}
	f_{s4} = \frac{\displaystyle \sum_{i=0}^{n}B_{pz}(i)}{n}
\end{equation}
where, $B_{pz}(i)$ represents the z-axis height corresponding to the Bezier curve.

\subsubsection{Adhesion Trajectory Path Length Constraint Strategy}
Given that adhesion trajectories are crucial for maintaining adhesion performance, minimizing the path length reduces energy consumption and avoids unnecessary travel.
\begin{equation} \label{eqa:adhesion_traj}
	f_{s5} = \int_{T_{2}}^{T_{3}} \sqrt{{(T_{t} B_{vx}(t))}^2 + {(T_{t} B_{vy}(t))}^2 + {(T_{t} B_{vz}(t))}^2 } dt   
\end{equation}
where $B_{vx}, B_{vy}, B_{vz}$ represents Bezier speed adhesion trajectories, $T_{t}$ represents the interval time.

\subsubsection{Trajectory Bending Constraint Strategy}
\begin{figure}[h]   
    \centering
    \includegraphics[width=0.4\textwidth]{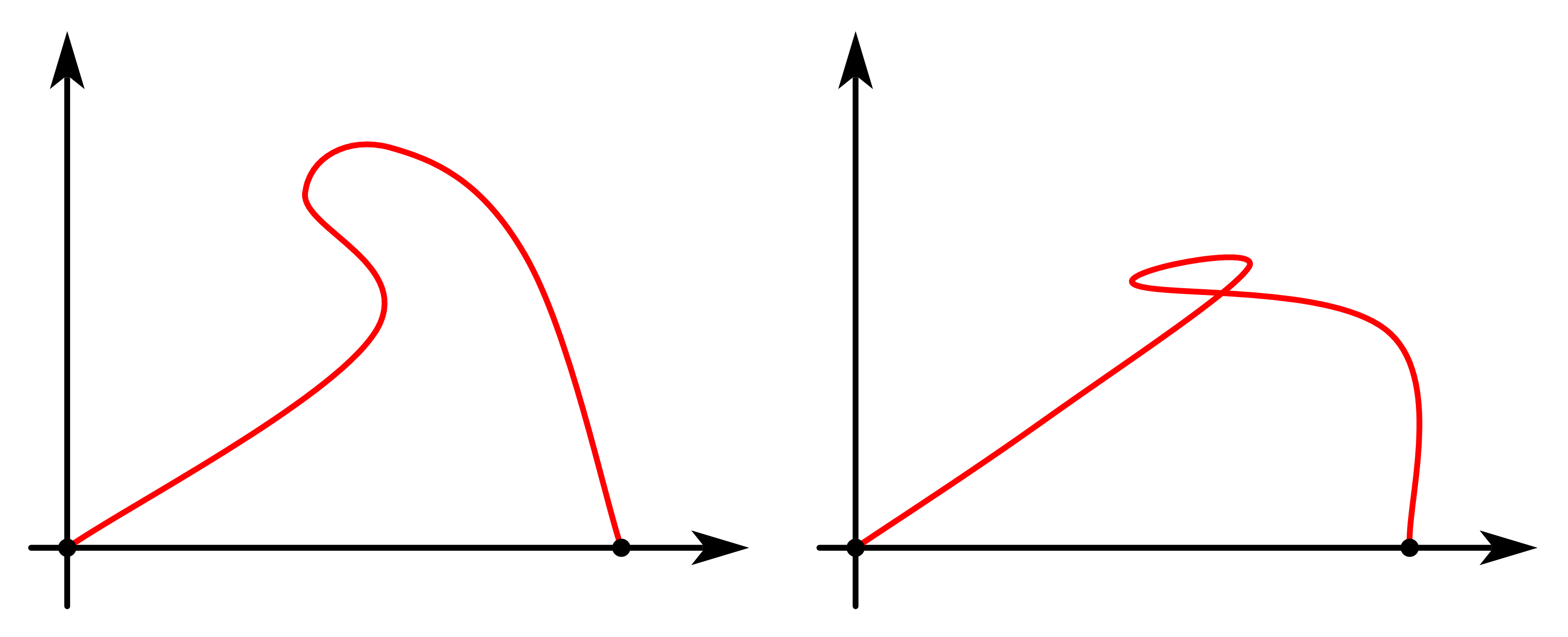}
    \caption{Schematic diagram of the bending phenomenon of Bezier curves.}
    \label{fig:trajectory_bending}
\end{figure}

This strategy aims to avoid the bending phenomenon of the optimal foot trajectory and ensure the smoothness of the trajectory, the trajectory bending form is shown in Fig. \ref{fig:trajectory_bending}. In this paper, we adopt a judgment method based on the angle between adjacent vectors. Specifically, the smoothness of the trajectory is evaluated by calculating the angle between neighboring vectors on the trajectory. If the angle between neighboring vectors exceeds a set threshold 90 °, the trajectory segment is considered to be curved. 

To quantify this bending phenomenon, this paper introduces the ${\rm ReLU}$ activation function, which utilizes its nonlinear characteristics to determine whether the angle exceeds the threshold, and sums up all the bending trajectory segments through the summation formula, to obtain a quantitative index of the degree of bending. The specific constraint form is as follows
\begin{equation} \label{eqa:bend_traj}
	f_{s6} = \displaystyle \sum_{t=0}^{T_{3}} {\rm ReLU}(\arccos(\frac{ \Delta \mathord{ \buildrel{ \lower3pt \hbox{$ \scriptscriptstyle \rightharpoonup$}} \over P_{t}} \Delta \mathord{ \buildrel{ \lower3pt \hbox{$ \scriptscriptstyle \rightharpoonup$}} \over P_{t}}}{| \Delta \mathord{ \buildrel{ \lower3pt \hbox{$ \scriptscriptstyle \rightharpoonup$}} \over P_{t}}| | \Delta \mathord{ \buildrel{ \lower3pt \hbox{$ \scriptscriptstyle \rightharpoonup$}} \over P_{t+1}}|}) - \frac{\pi}{2})
\end{equation}
where $\Delta \mathord{ \buildrel{ \lower3pt \hbox{$ \scriptscriptstyle \rightharpoonup$}} \over P_{t}}$ is the vector value of the neighboring trajectory point at the foot position at the current moment; and $\Delta \mathord{ \buildrel{ \lower3pt \hbox{$ \scriptscriptstyle \rightharpoonup$}} \over P_{t+1}}$ is the vector value of the neighboring foot end position trajectory point at the next moment.

\subsubsection{Foot Jitter Constraint Strategy}

This strategy aims to constrain the jitter of the foot during the detachment phase. To quantify the jitter, the second-order derivative of the detachment force trajectory is used to represent it, i.e., the jitter characteristics are captured by analyzing the rate of change of the force. Further, the z-score rule is applied to identify abnormal jitter points. The ${\rm ReLU}$ activation function removes jitter points smaller than three times the standard deviation, leaving only large jitter points that exceed three times the standard deviation, and finally accumulates all jitter points. This processing can effectively suppress abnormal jitter, thus improving the stability of the robot in the detachment process. The specific constraint form is as follows
\begin{equation} \label{eqa:shake_force}
	f_{s7} = \displaystyle \sum_{i=0}^{n} {\rm ReLU}(|\frac{[{F_{d}}''(B_p)]_{i} - \mu}{\sigma}| - m)
\end{equation}
where ${F_c}''(B_p(t))$ is the second-order derivative of the foot-end detachment force trajectory, $\mu$ is the mean value of the second-order derivative of the foot-end detachment force trajectory, $\sigma$ is the standard deviation of the second-order derivative of the foot-end detachment force trajectory, and $m$ denotes more than a factor of three standard deviation.

\subsection{The Pareto Front of the Non-dominated Solution}
To find the Pareto frontiers containing non-dominated solutions for all optimal foot trajectories according to different climbing tasks, multi-objective optimization can be achieved by employing any number of the above-mentioned strategy models as cost functions and constraints, and by optimizing the Bessel control point $P_{3x}$, $P_{4x}$, $P_{5x}$, $P_{10x}$, $P_{11x}$, $P_{12x}$, $P_{3z}$, $P_{4z}$, $P_{5z}$, $P_{10z}$, $P_{11z}$ and $P_{12z}$. For this purpose, this paper uses the NSGA-II\cite{NSGAII} algorithm for multi-objective optimization. This algorithm is a fast elite multi-objective genetic algorithm, which can solve the drawbacks of the current Nondominated Sorting Genetic Algorithm, such as too high time complexity of nondominated sorting, lack of elite retention strategy, and the need to specify a shared parameter to ensure the diversity of the population, and it can quickly find a better distribution of the solutions and better convergence of Pareto's method for multi-objective optimization. convergence of the Pareto frontiers.

\subsection{Redundancy Hierarchical Strategy}
Traditional methods usually use weights assigned to find the optimal solution from the Pareto frontier. However, since the weights of this method are given artificially and not obtained in a standard way, there are some limitations.

To solve this problem, this paper proposes a redundant hierarchical strategy. This strategy ensures that important tasks are prioritized to be solved by ranking the task objectives and adding redundancy factors to the objectives at each level. The redundancy factor is used to control the priority of the objectives at different levels, and the lower the level, the smaller the redundancy factor, thus gradually narrowing the scope of the optimal solution. In this way, the solution that best meets the actual task requirements can be filtered out of the Pareto frontier that contains all the optimal foot end trajectory non-dominated solutions. The core idea of this strategy is to prioritize the solutions that satisfy the key performance indicators according to the specific requirements of the task, while taking into account the balance of other constraints, to ensure that the final trajectory selected is the optimal trajectory that meets the requirements. The specific algorithm of the redundancy hierarchical strategy is shown in Algorithm \ref{alg:RHS}, which includes the following five steps.

\begin{algorithm}[h]
	\small
	\caption{Redundancy Hierarchical Strategy Algorithm}
	\label{alg:RHS}
	\begin{algorithmic}
		\Require Pareto front $P_{f}$
		\Ensure Optimal Bezier Trajectory $B_{o}$
		\State initialize $\alpha \in (0,1)$, $\beta \in (0,1)$
		\State get $N_{c}$, $N_{r}$, $M$;
		\Repeat 
		\State calculate $V_{mi}$, $V_{ma}$;
		\State calculate $V_{r} \gets V_{mi} + (V_{ma} - V_{mi}) \alpha$; \Comment{update $V_r$}
		\State update $\alpha = \alpha  \beta^{i}$;
		\If{$P_{f}[j][i] > V_{r}$}
		\State removal the corresponding element $j$ in the vector $M$;
		\EndIf
		\Until{$\mathrm{len(M)} = 1$}  \Comment{termination conditions}
		\State Obtaining optimal foot trajectory $B_{o}$
	\end{algorithmic}
\end{algorithm}

1) Initialize the redundancy factor and decay coefficient: first, initialize the values of the redundancy factor $\alpha \in (0,1)$  and decay coefficient $\beta \in (0, 1)$. The redundancy factor $\alpha$ is used to adjust the weights of the objectives at each level, while the decay coefficient $\beta$ controls the rate of change of the redundancy factor during the iteration process.

2) Calculate the dimensions of the Pareto frontier: With the acquired Pareto frontier $P_f$, calculate its number of columns $N_c$ and rows $N_r$ and construct the vector based on these dimensions $M$. $M$ is a vector from 0 to $N_r$ with step size 1.

3) Cyclic Iterative Optimization: in each iteration, the minimum value of $V_{mi}$ and the maximum value of $V_{ma}$ in column $i$ of the Pareto frontier $P_f$ are computed based on the number of loops $i$. then, by using Eq. $V_{r} \gets V_{mi} + (V_{ma} - V_{mi}) \alpha$ Calculate the judgment value $V_r$ and update the redundancy factor $\alpha = \alpha  \beta^{i}$. 

4) Objective Judgment and Removal: the data in column $i$ in the Pareto frontier $P_f$ is judged. If the element in row $j$ in that column is larger than the current judgment value $V_r$, then the corresponding element $j$ in vector $M$ is removed.This process gradually reduces the optimization space by progressively eliminating solutions that do not meet the redundancy requirements.

5) Termination condition and optimal trajectory selection: the algorithm terminates when there is only one element left in the vector $M$. At this point, the objective function value corresponding to the remaining element in $M$ is the optimal objective function. Based on the solution of this objective function, the corresponding Bessel control points are computed, and the optimal foot trajectory $B_o$ is finally obtained.
\section{FTFOF Algorithm Performance Verification Experiment}
\label{sec:Experimental Results and Analysis}

In order to validate the performance of the Foot Trajectory and Force Optimization Framework (FTFOF) algorithm, this paper chooses the most commonly used straight line climbing mode as the entry point for validation, and uses the laboratory-developed MST-M3F quadruped climbing robot as the carrier. The straight line climbing mode is simple and representative, and is an ideal scenario for evaluating the merits of the robot's foot-end trajectories, which allows for an intuitive comparison of the performance differences between different foot-end trajectories.

It is worth noting that the optimization framework proposed in this paper is not only limited to the straight line climbing mode, but also applicable to other kinds of climbing modes and task scenarios. When switching between climbing modes or tasks, the process of generating optimal foot trajectories is very simple. Specifically, the mobile generalized data acquisition platform is used to collect training samples corresponding to different climbing modes and tasks, and then train the foot trajectory and detachment force model, as well as the foot trajectory and pre-pressure model based on these data.

\subsection{Foot Climbable Trajectory Set}
For the user, the acquisition of the set of climbable trajectories at the foot can be accomplished by simply obtaining five constraints based on a specific foot structure in the same way as the foot structure constraint policy described above.

\subsubsection{Maximum Foot Position Constraints}
According to Eq. \ref{eqa:foot_pos_limit} the final optimal solution for the obtained joint angles can be obtained as $[\theta_{1}, \theta_{2}, \theta_{3}]=[0.660, 2.191, -2.467]$ (rad), corresponding to a maximum position limit of $P_{pl} =(P_{plx}, P_{plz}) = (0.189, 0.074)$ (m). To ensure safe motion, a safety factor of 85 $\%$ was set, and the final maximum position boundary was obtained as $P_{pl} = (0.160, 0.064)$ (m).For better comparison with other commonly used trajectories, the foot landing point was set to be $P_{15} = (0.12,0) $ (m).

\subsubsection{Maximum Foot Velocity Constraints}
According to Eq. \ref{eqa:foot_vel_limit}, the optimal solution of the optimized variables, i.e., the optimal solution of the joint angle and the joint angular velocity is $[\theta_{1}, \theta_{2}, \theta_{3}, \dot{\theta_{1}}, \dot{\theta_{2}}, \dot{\theta_{3}}]=[-0.506, 1.573, -2.412, 7.3, 6.8, 6.8]$ (rad, rad/s), and the corresponding maximal velocity limit is $V_{pl} = (V_{plx}, V_{plz}) = (1.012, 0.961)$ (m/s), with a fusion safety factor of 80 $\%$, and a final maximum velocity bound of $V_{pl} = (0.80, 0.77)$ (m/s).

\subsubsection{The Minimum Detachment Point Constraint}
The robot's foot structure consisted of a rigid foot and flexible adhesive material with a PET film, which cannot detach directly in any direction due to its rigidity. Testing indicated a minimum bend radius of $f_{c} = 6$ mm without permanent deformation. To ensure no damage to the PET film, the minimum desorption height was set to $h_{d} = 1.5*f_{c}$.

\begin{figure}[h]   
    \centering
    \includegraphics[width=0.3\textwidth]{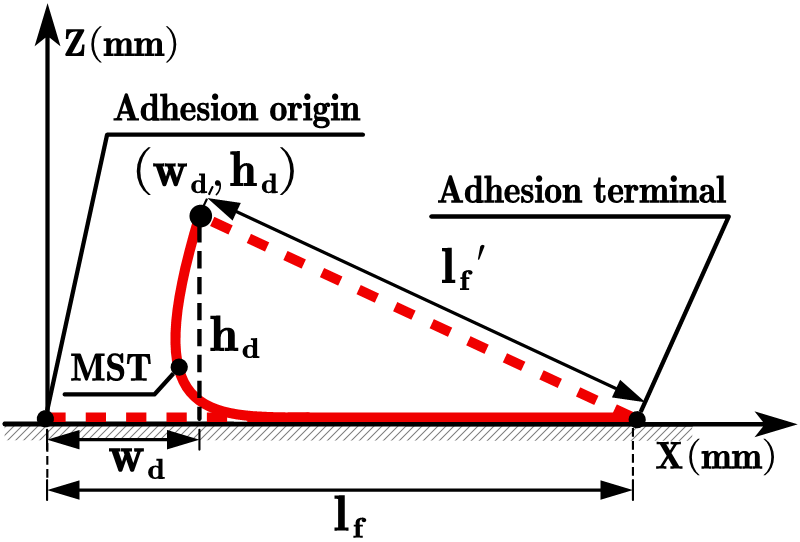}
    \caption{Maximum Detachment Point Constraints for Foot.}
    \label{fig:minimum_detachment_point_constraint}
\end{figure}

In addition, since it is a linear motion and there is no displacement in the $y$ direction, there is no need to consider $w_d$. Specifically, as shown in the Fig. \ref{fig:minimum_detachment_point_constraint}. The adhesive material length was $l_{f}=50$ mm. Assuming a maximum detachment length $l_{d} = l_f - \sqrt{l_{f}^2 - h_{d}^2} = 2.6$ mm, $l_{d}$ was set at 4.5 mm, a median value between calculations. The final minimum detachment point obtained $m_d = (0.009,0.0045)$ (m).

\subsubsection{Transition Control Point Constraints}
Control point $P_{5}$ was set to be greater than the minimum detachment point but within the foot trajectory position boundary. 
\begin{equation} \label{eqa:P5_limits}
	\begin{cases}
        l_{d} < P_{5x} < P_{lx} \\
		h_{d} < P_{5z} < P_{lz} 				
	\end{cases}
\end{equation}

\begin{figure}[H]   
    \centering
    \includegraphics[width=0.3\textwidth]{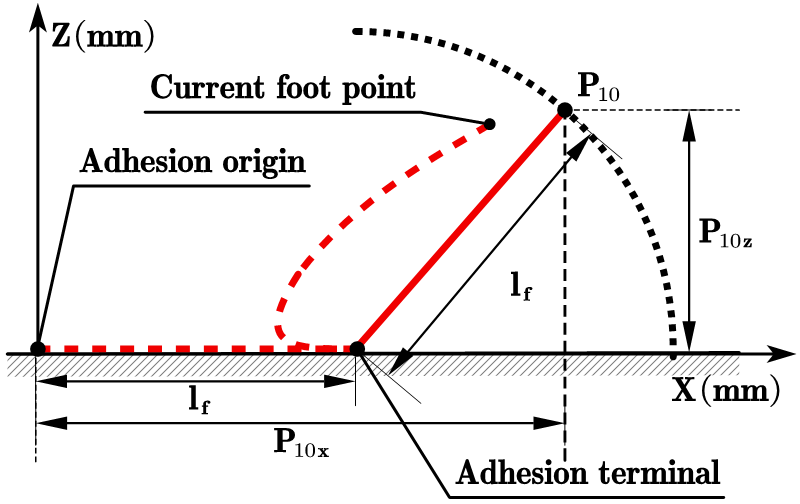}
    \caption{Schematic Diagram of Control Point $P_{10}$ Constraints.}
    \label{fig:P_10_constraint}
\end{figure}

For $P_{10x}$, it is crucial to ensure that $P_{10x}$ is greater than the length of the adhesive material, $l_f$. If $P_{10x}$ is shorter than $l_f$, the overall detachment height will increase, leading to a greater detachment angle of the adhesive material and consequently requiring additional detachment force. Additionally, constraints on $P_{10z}$ must be considered, as illustrated in Fig. \ref{fig:P_10_constraint}. The red dashed line in the figure represents the adhesion material's detachment shape during the process. Considering the limiting case, to guarantee complete detachment, the distance between the end of the adhesive material and the point $P_{10}$ must equal $l_f$. This results in $P_{10z} = \sqrt{l_{f}^2 - (P_{10x} - l_{f})^2}$. Furthermore, due to the droop of the PET film used on the foot, $P_{10z}$ must exceed the droop height $h_{s}$. Moreover, along the x-axis, the endpoint of the swing trajectory must be greater than the starting point, ensuring $P_{10x} > P_{5x}$.

\begin{equation} \label{eqa:P10_limits}
	\begin{cases}
		\max(P_{5x},l_{f}) < P_{10x} < P_{plx} \\ 
		\max(h_{s},(\sqrt{l_f^2 - (P_{10x} - l_f)^2})) < P_{10z} < P_{plz}
	\end{cases}
\end{equation}

\subsubsection{Trajectory Shape Constraints}
To ensure that the PET film did not suffer permanent deformation and that stable detachment was achieved, control points $P_{3}, P_{4}$ were restricted to the upper triangle formed by the detachment trajectory's start and end points $P_{0}, P_{5}$. 
\begin{equation} \label{eqa:detachment_limits2}
	\begin{cases}
		P_{0x} < P_{3x} < P_{5x} \\
		P_{3x} \cdot \bigtriangleup < P_{3z} < P_{5z} \\
		P_{3x} < P_{4x} < P_{5x} \\
		P_{3z} \cdot \bigtriangleup < P_{4z} < P_{5z} \\
	\end{cases}
\end{equation}
where $\bigtriangleup$ denotes the slope between control points $P_{0}$ and $P_{5}$.

For the adhesion trajectory, the convex hull property was maintained to prevent the adhesive material from folding upon landing, optimizing adhesion efficiency. An arc adhesion strategy was adopted, ensuring that control points $P_{11}, P_{12}$ were positioned to the right of the landing point.

\begin{equation} \label{eqa:arc_adhesion_limit}
	\begin{cases}
		P_{15x} < P_{11x} < P_{plx} \\
		P_{15z} < P_{11z} < P_{plz} \\
		P_{15x} < P_{12x} < P_{plx} \\
		P_{15z} < P_{12z} < P_{plz} \\
	\end{cases}
\end{equation}

The swing trajectory was a transition trajectory, requiring control points$P_6, P_7, P_8, P_9$ to be within the quadrilateral formed by the limits $P_{0}, P_{l}$. 
\begin{equation} \label{eqa:swing_points}
	\begin{cases}
		P_{0} < P_{6} < P_{pl} \\
		P_{0} < P_{7} < P_{pl} \\
		P_{0} < P_{8} < P_{pl} \\
		P_{0} < P_{9} < P_{pl} \\
	\end{cases}
\end{equation}

By combining Eqs. \ref{eqa:c2_continuous_constraint_point}, \ref{eqa:detachment_limits2} and \ref{eqa:arc_adhesion_limit},  the final constraints for the control points $P_{3}, P_{4}, P_{11}$ and $P_{12}$ are obtained as follows.
\begin{equation} \label{eqa:P_3_4_11_12_final}
	\begin{cases}
		P_{4xl} < P_{4x} < P_{4xr} \\
        P_{4zl} < P_{4z} < P_{4zr} \\

        P_{3xl} < P_{3x} < P_{3xr} \\
        P_{3zl} < P_{3z} < P_{4zr} \\

        P_{11xl} < P_{11x} < P_{11xr} \\
        P_{11zl} < P_{11z} < P_{11zr} \\

        P_{12xl} < P_{12x} < P_{12xr} \\
        P_{12zl} < P_{12z} < P_{12zr} \\
	\end{cases}
\end{equation}

By fusing the above constraints with universal foot trajectories and by transforming the control point positions, the set of foot climbable trajectories is obtained.

\subsection{Generalized Data Acquisition and Model Training}
A random sampling method was employed to select 1000 sample data points from the total sample of climbable foot trajectories. Force data corresponding to these samples was collected using a force acquisition platform, resulting in a unique set of data samples. 

\begin{figure}[h]
\centering  
\subfigure[]{
    \includegraphics[width=0.45\linewidth]{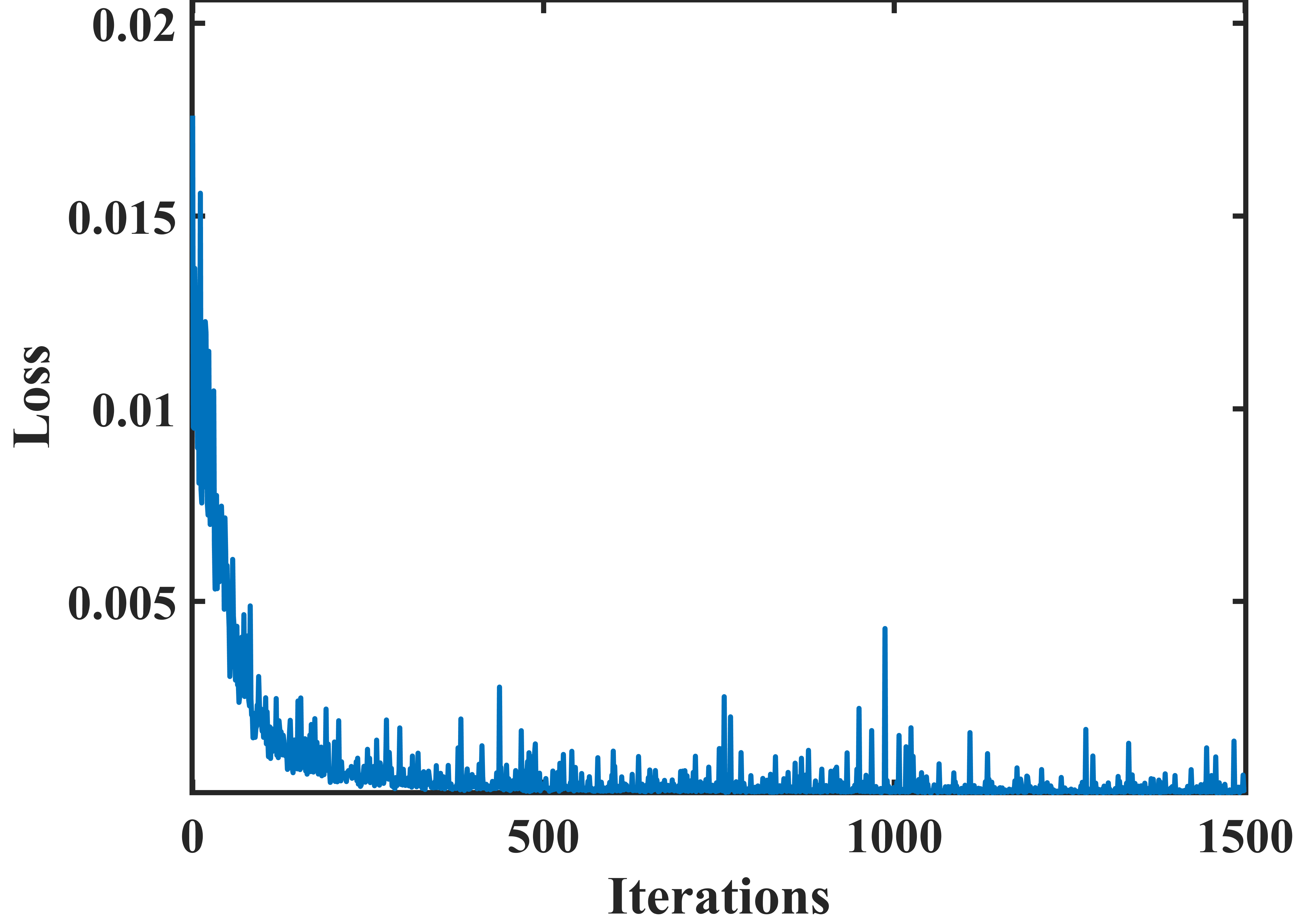}}
\subfigure[]{
    \includegraphics[width=0.45\linewidth]{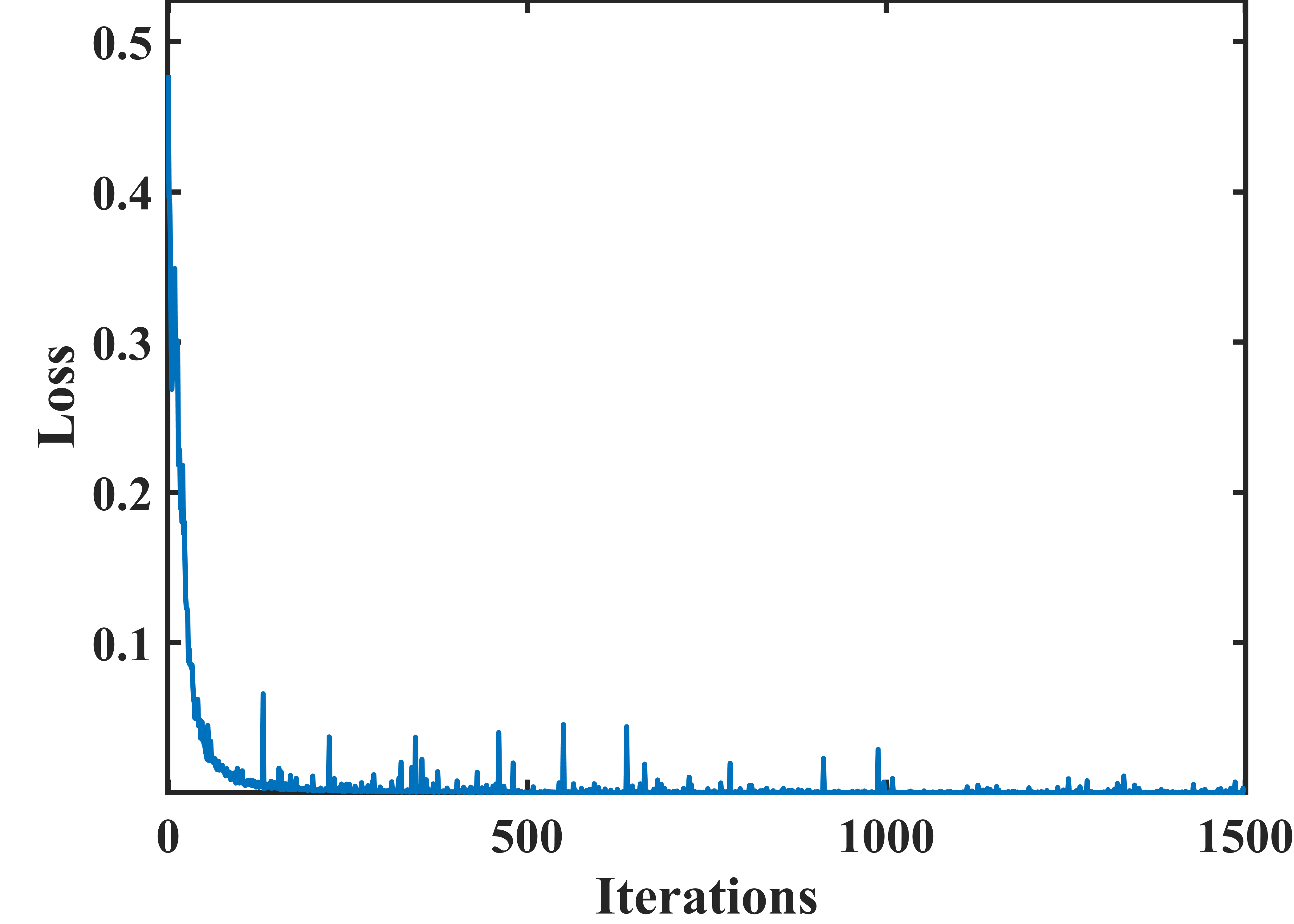}}
\caption{Validation set loss curve of model training process. (a) Pre-pressure verification set loss diagram; (b) Loss diagram of desorption verification set.}
    \label{fig:loss_adhesion_detachment}
\end{figure}

The collected sample data are input into the GRU model based on the DILATE loss function for training to obtain the foot trajectory-detachment force model $F_d$ and the foot trajectory-prepressure model $F_p$. In the process of model training, to ensure the model's generalization ability and accuracy, the change of the validation loss needs to be closely attended to, and the graph of the validation set loss in the training process is shown in Fig. \ref{fig:loss_adhesion_detachment}. Fig. \ref{fig:loss_adhesion_detachment} (a) is the prepressure validation set loss plot, which is the adhesion force loss value of the validation set, and Fig. \ref{fig:loss_adhesion_detachment} (b) is the detachment force validation set loss plot. This figure demonstrates the improvement in model performance during training. As the training progresses, the validation loss gradually decreases, indicating that the model learns well from the sample data and can accurately capture the complex relationship between the foot trajectory and the force, and eventually the validation loss stabilizes. 

\begin{figure}[!htbp]
\centering  
\subfigure[]{
    \includegraphics[width=0.46\linewidth]{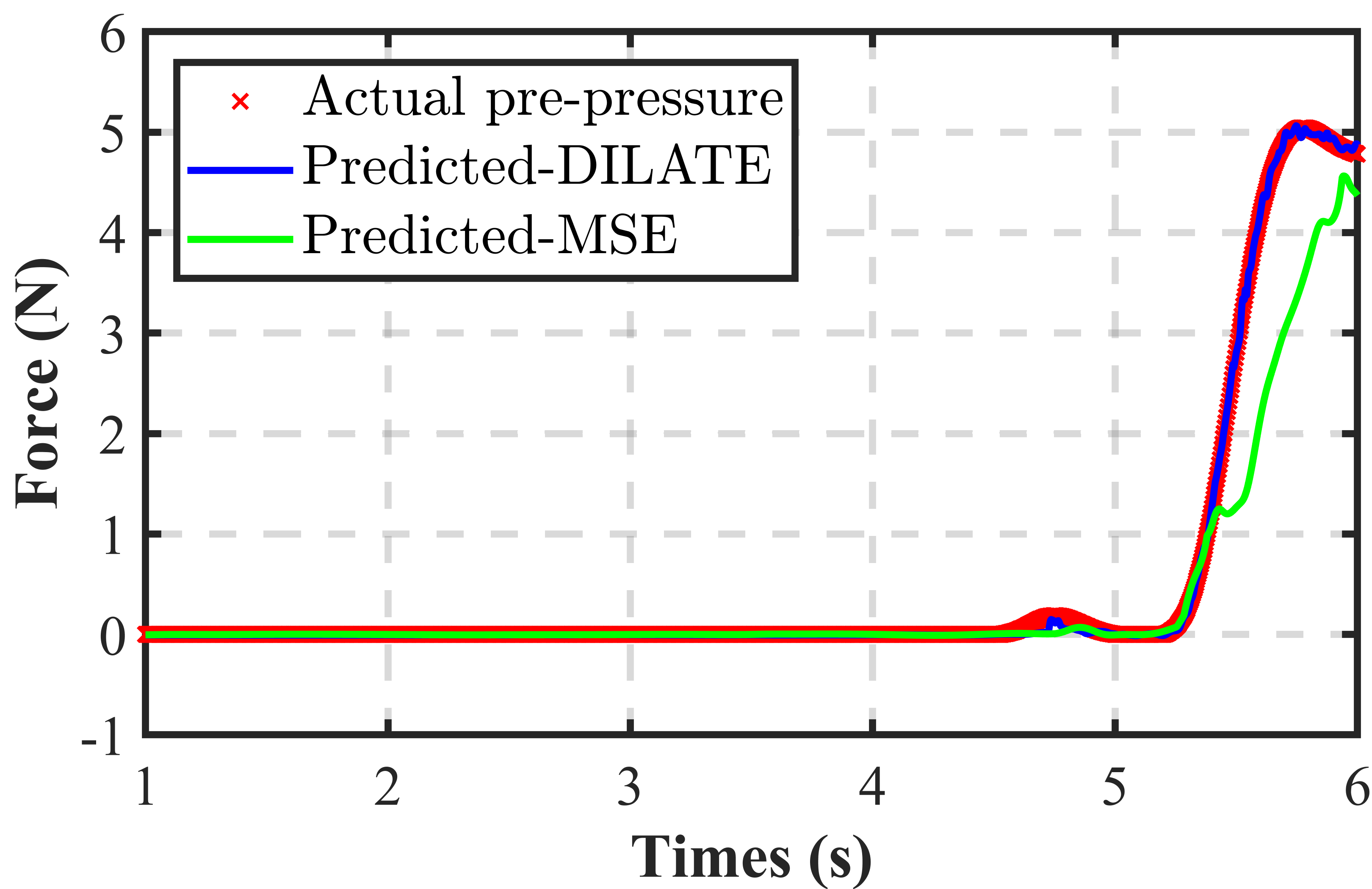}}
\subfigure[]{
    \includegraphics[width=0.46\linewidth]{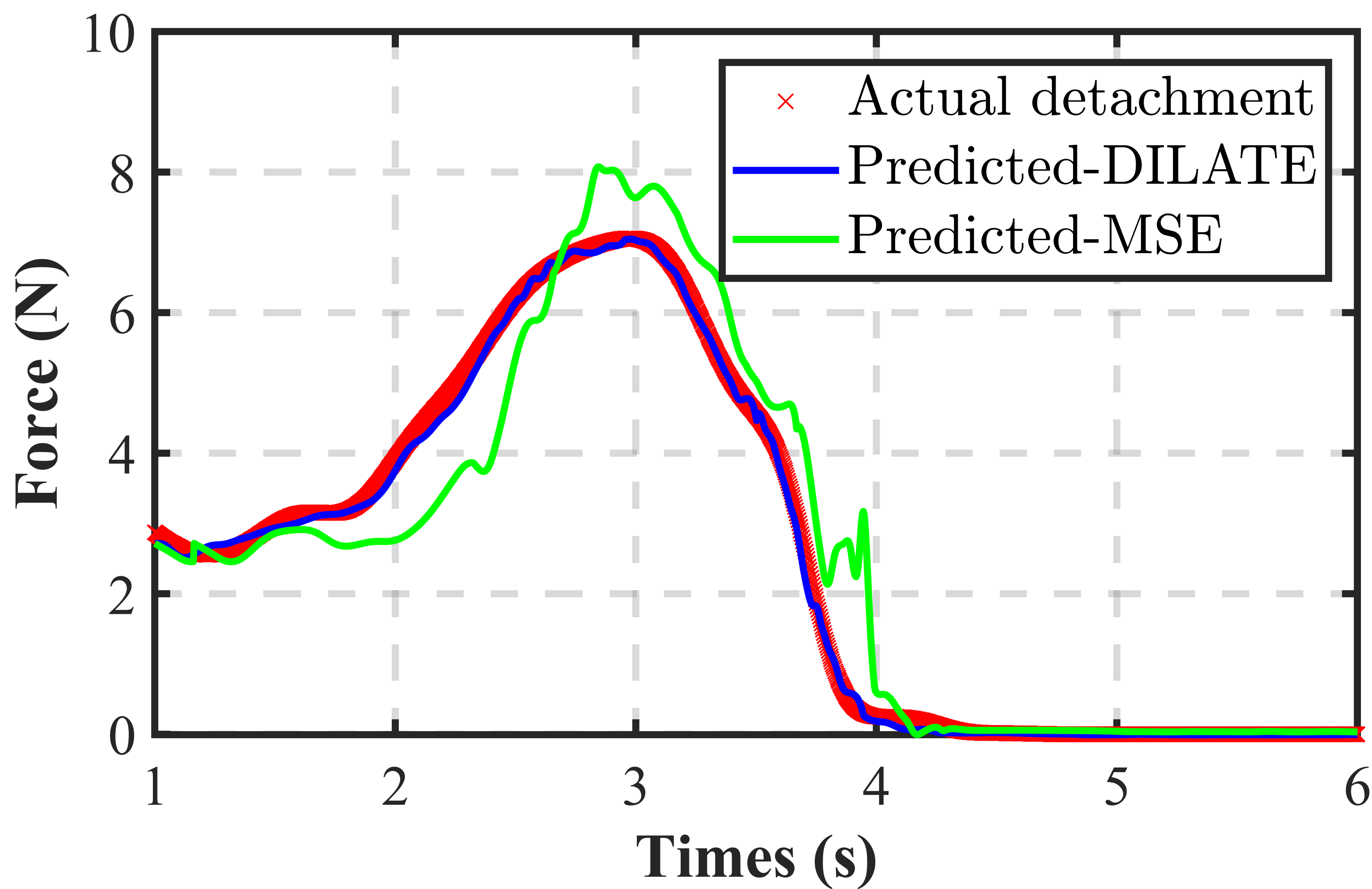}}
\caption{Comparison of GRU model prediction data with validation set data. (a) Comparison of pre-pressure data; (b) Comparison of detachment force data.}
    \label{fig:forecast_real_force_compare}
\end{figure}

To further verify the advantages of the DILATE loss function over the traditional MSE (Mean Square Error) loss function, this paper trained the MSE-based GRU model and the DILATE-based GRU model using the same sample data set, and compared the performance of the two models in detail. The specific comparison results are shown in Fig. \ref{fig:forecast_real_force_compare}.

This comparative experiment clearly demonstrates the advantages of the DILATE loss function in dealing with time-series data, especially in scenarios involving shape and time distortions. The DILATE loss function can deal with dynamic changes in the data more efficiently by introducing a shape and time alignment mechanism, which improves the predictive performance of the model.

\subsection{Multi-Objective Optimization}

The trajectory bending constraint strategy $f_{s6}$ is chosen as an equation constraint, and the maximum detachment force constraint strategy $f_{s1}$, the adhesion performance constraint strategy $f_{s3}$, the foot jitter constraint strategy $f_{s7}$, the average detachment force constraint strategy $f_{s2}$, and the adherence trajectory route length constraint strategy $f_{s5}$ are used as objective functions for the solution of the Pareto front. The specific objective functions and constraint equations are as follows
\begin{equation} \label{eqa:multi_objective_fun}
    \begin{aligned}
        &\ \min F_{s}(P) = (f_{s1}(P), f_{s3}(P), f_{s7}(P), f_{s2}(P),f_{s5}(P)) \\
        &\ \text { s.t. } 
			\begin{cases}
			& f_{s6} = 0 \\
			& \min P <  P < \max P \\
			& m_{d} <  B_{p} < P_{pl} \\
			& -V_{pl} <  B_{v} < V_{pl}
		  	\end{cases}
    \end{aligned}
\end{equation}
where $B_{p}$ and $B_{v}$ represent the foot position trajectory and velocity trajectory, respectively, which need to be qualified by the maximum foot position constraint, minimum detachment point constraint, and maximum foot velocity constraint. $P$ represents all control point optimization variables: i.e., $P_{3x}$, $P_{4x}$, $P_{5x}$, $P_{10x}$, $P_{11x}$, $P_{12x}$, $P_{3z}$, $P_{4z}$, $P_{5z}$, $P_{10z}$, $P_{11z}$ and $P_{12z}$. And $\min P < P < \max P$ denotes the constraints for the control points, which need to satisfy the above transition control point constraints and trajectory shape constraints, i.e., the constraints of Eqs. (\ref{eqa:P5_limits}), Eqs. (\ref{eqa:P10_limits}), and Eqs. (\ref{eqa:P_3_4_11_12_final} ) constraints.

\begin{figure}[h]   
    \centering
    \includegraphics[width=0.4\textwidth]{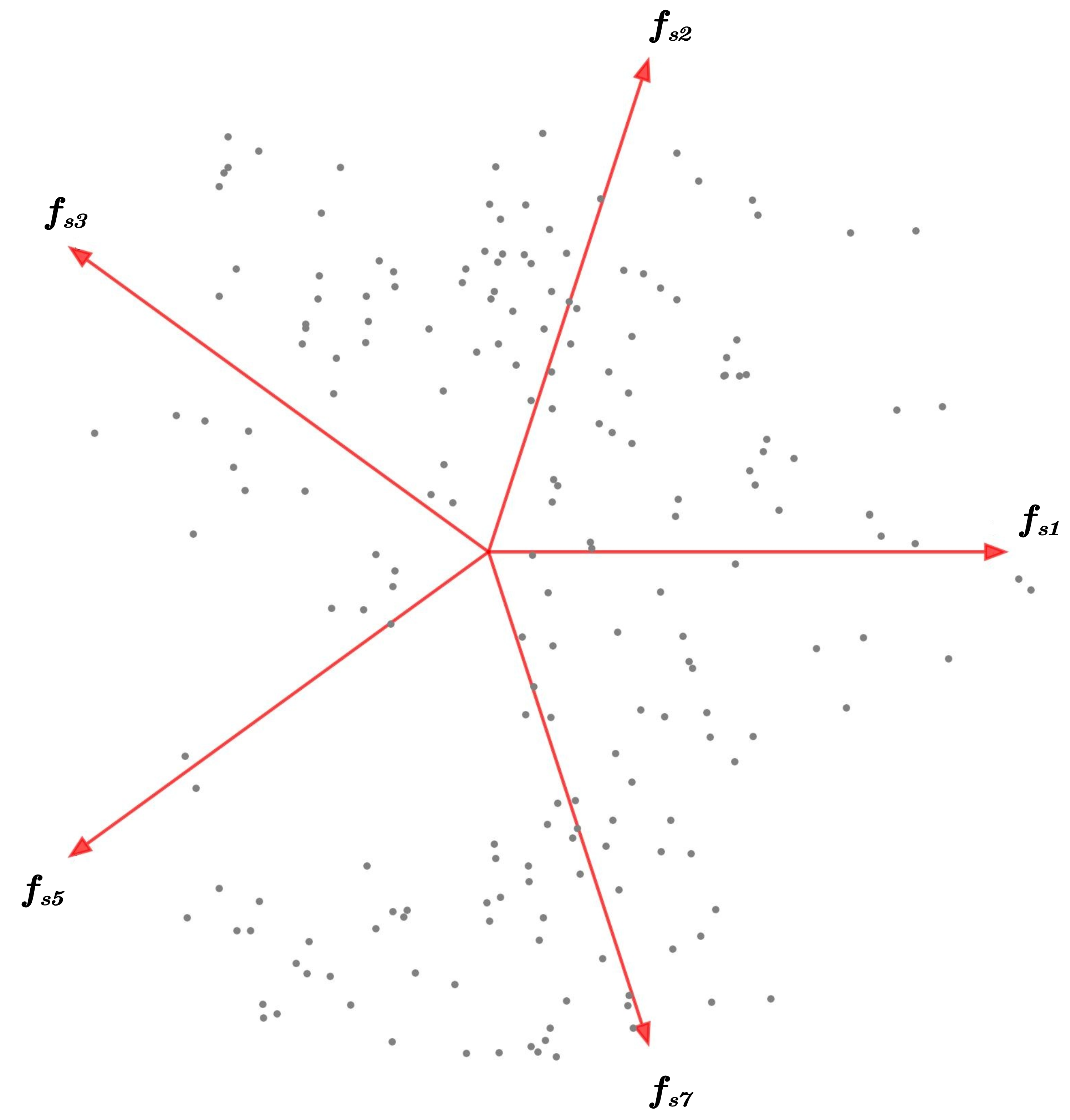}
	\caption{Pareto front results in the target space of the foot trajectory are presented}
    \label{fig:Star_Coordinate_Plot}
\end{figure}

In order to solve the above mentioned multi-objective optimization problem, the pymoo\cite{pymoo} multi-objective optimization solver is used as a solution tool, and the NSGA2\cite{NSGAII} function is chosen as a multi-objective optimization algorithm. In the optimization process, the population size was set to 200, and the maximum number of iterations was 2000. Eventually, the Pareto front containing non-dominated solutions for all foot trajectories can be obtained. Since the objective function exists in a high-dimensional space, it is difficult to display it directly, so this paper adopts the Star Coordinate method for visualization. The Pareto front in the objective space of the specific end-of-foot trajectory is shown in Fig. \ref{fig:Star_Coordinate_Plot}. Among them, the gray points represent the characterization of the Pareto front in the multidimensional target space in the two-dimensional space.

In order to filter the optimal trajectory that best meets the task requirements from the Pareto frontiers of the foot trajectories, the redundancy grading strategy \ref{alg:RHS} proposed in this paper is used to realize it. First, the objective function is sorted into $f_{s1}, f_{s3}, f_{s7}, f_{s2}, f_{s5}$. according to priority. and the initial redundancy factor $\alpha$ = 0.9 and the attenuation coefficient $\beta$ = 0.9. With this strategy, the optimal solution that meets the requirement is finally found in the Pareto front $P_f$ =[2.3, 18.3, 31.4, 108.8, 123.7, 124.5, 23.6,25.5, 29.0, 48.6, 30.5, 11.1], and the optimal control point corresponding to the optimal foot trajectory is obtained.

\begin{figure}[h]   
    \centering
    \includegraphics[width=0.4\textwidth]{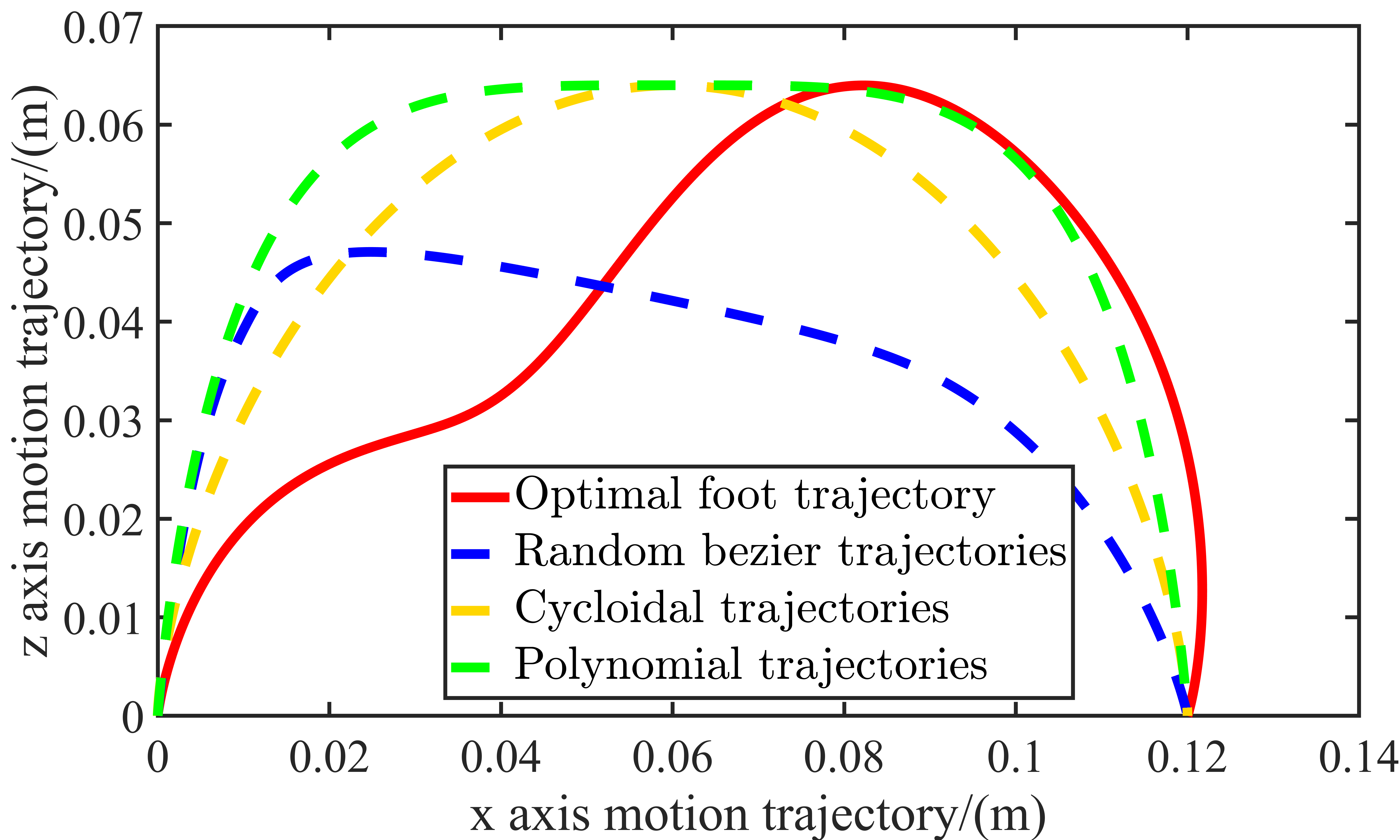}
	\caption{Comparison between optimal foot trajectory and common foot trajectory}
    \label{fig:compare_traj}
\end{figure}

To verify the performance advantages of the FTFOF proposed in this paper, the resulting optimal foot trajectories are compared with several trajectories commonly used in legged robots, which include random bezier, cycloidal and polynomial trajectories, as shown in Fig. \ref{fig:compare_traj}. Through this comparison, the differences in shape and characteristics of the different trajectories can be visualized.

\subsection{Results Analysis}
\begin{figure}[!htbp]
\centering  
\subfigure[]{
    \includegraphics[width=0.7\linewidth]{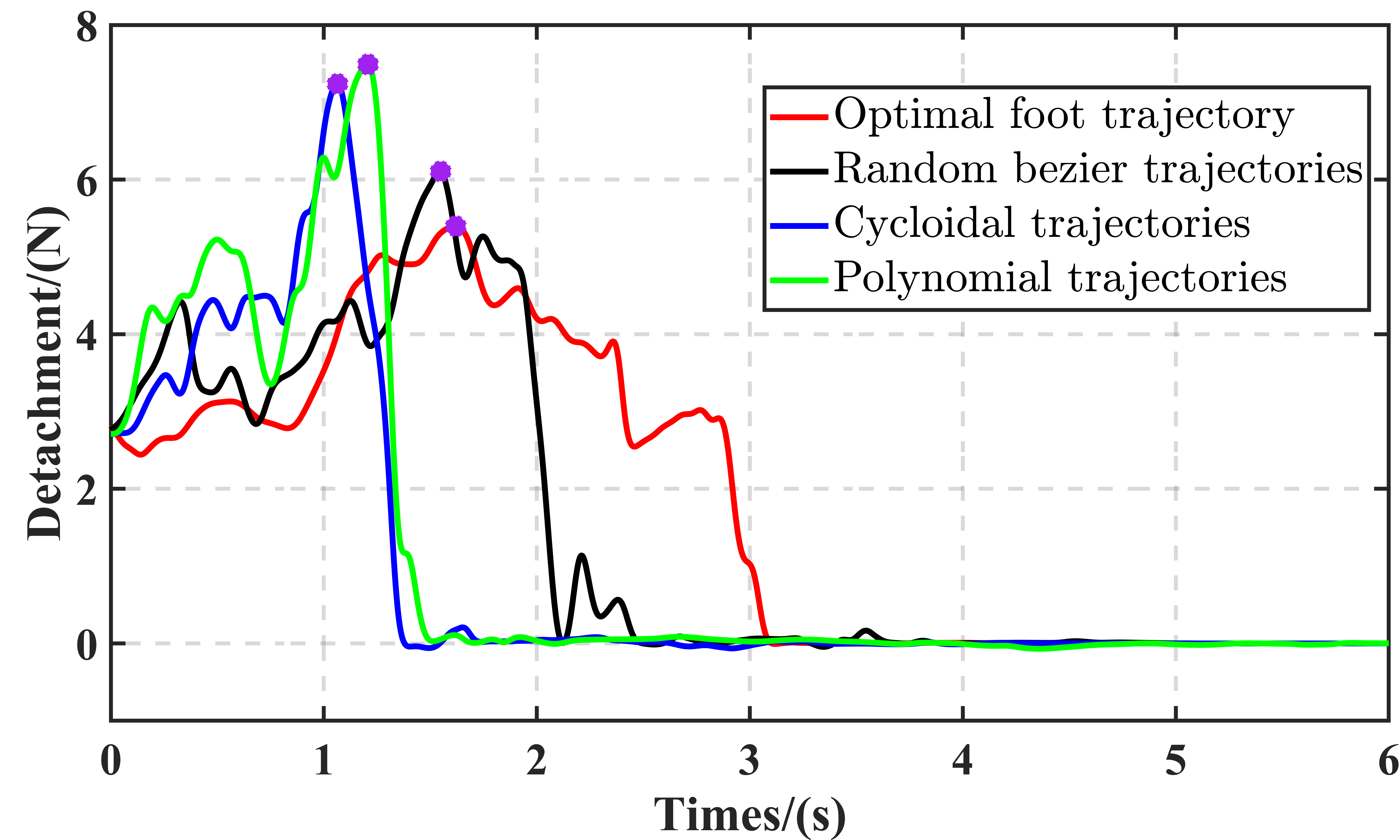}}
\subfigure[]{
    \includegraphics[width=0.7\linewidth]{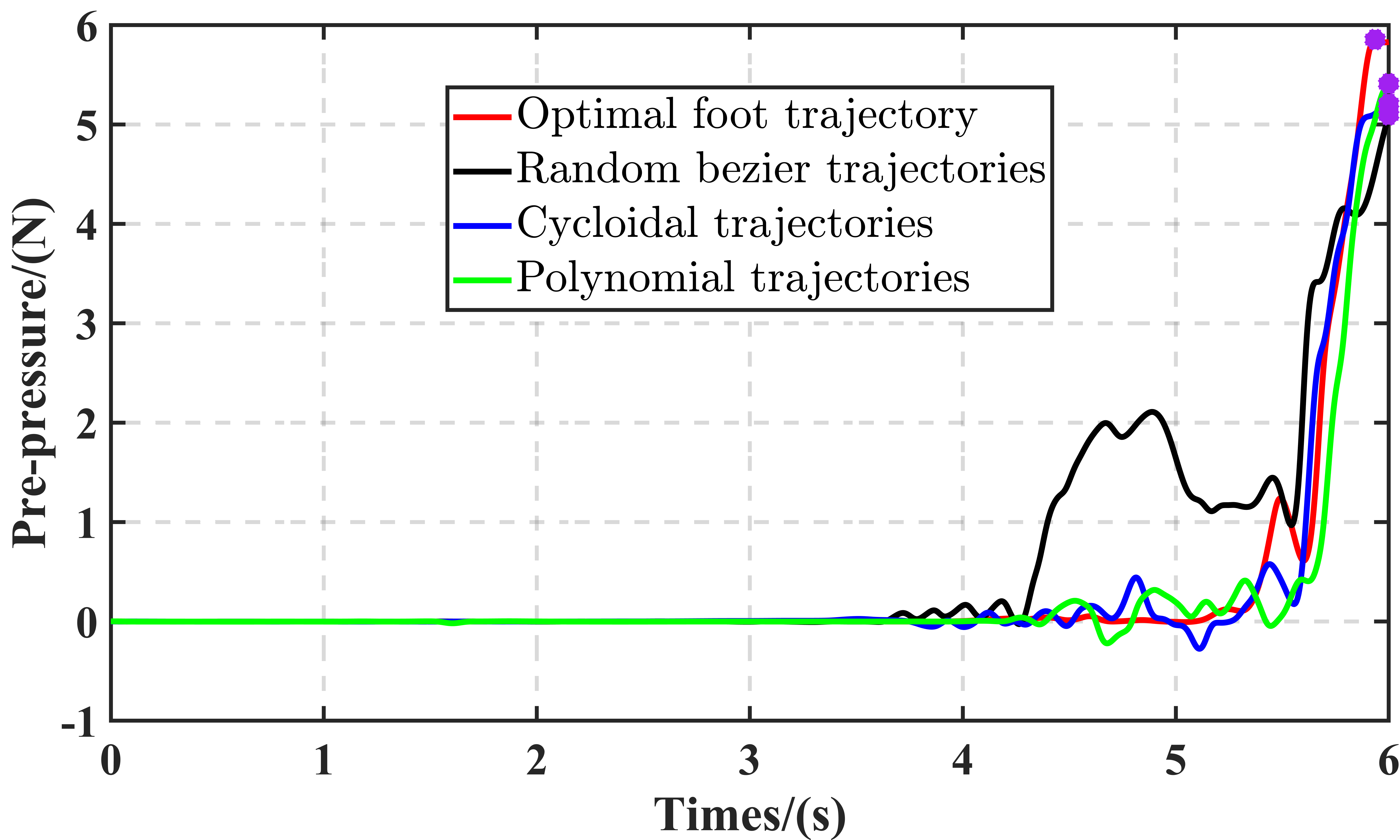}}
\caption{Comparison of foot force corresponding to optimal foot trajectory and common foot trajectory. (a) Comparison of foot detachment force; (b) Comparison of foot pre-pressure}
    \label{fig:diff_traj_adhesion_detachment_force_compare}
\end{figure}

In this paper, we compare and analyze two key aspects to highlight the advantages of the acquired optimal foot end trajectories. First, the optimal foot trajectory is compared with several trajectories commonly used in legged robots, using the robot's single-leg motion performance as a criterion. Specifically, by placing a single leg on a mobile force generalized data acquisition platform, the detachment force and preload force during the motion process are tested as the basis for judgment. The foot force comparison is shown in Fig. \ref{fig:diff_traj_adhesion_detachment_force_compare}.

\begin{figure}[h]   
	\centering
    \includegraphics[width=0.48\textwidth]{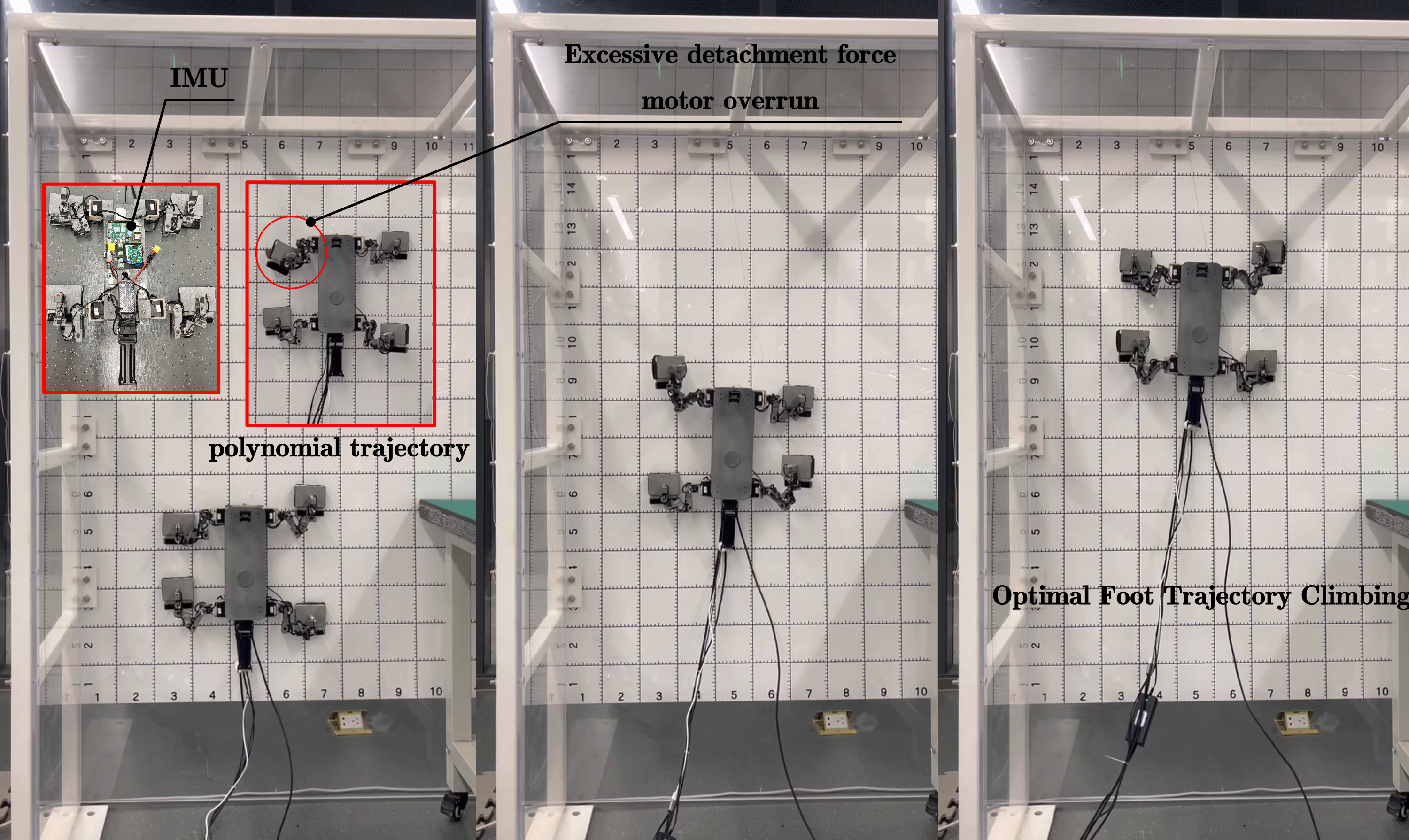}
	\caption{MST-M3F test body jitter data experiment}
	\label{fig:IMU_Vibration_climbing}
\end{figure}

\begin{figure}[h]   
	\centering
    \includegraphics[width=0.4\textwidth]{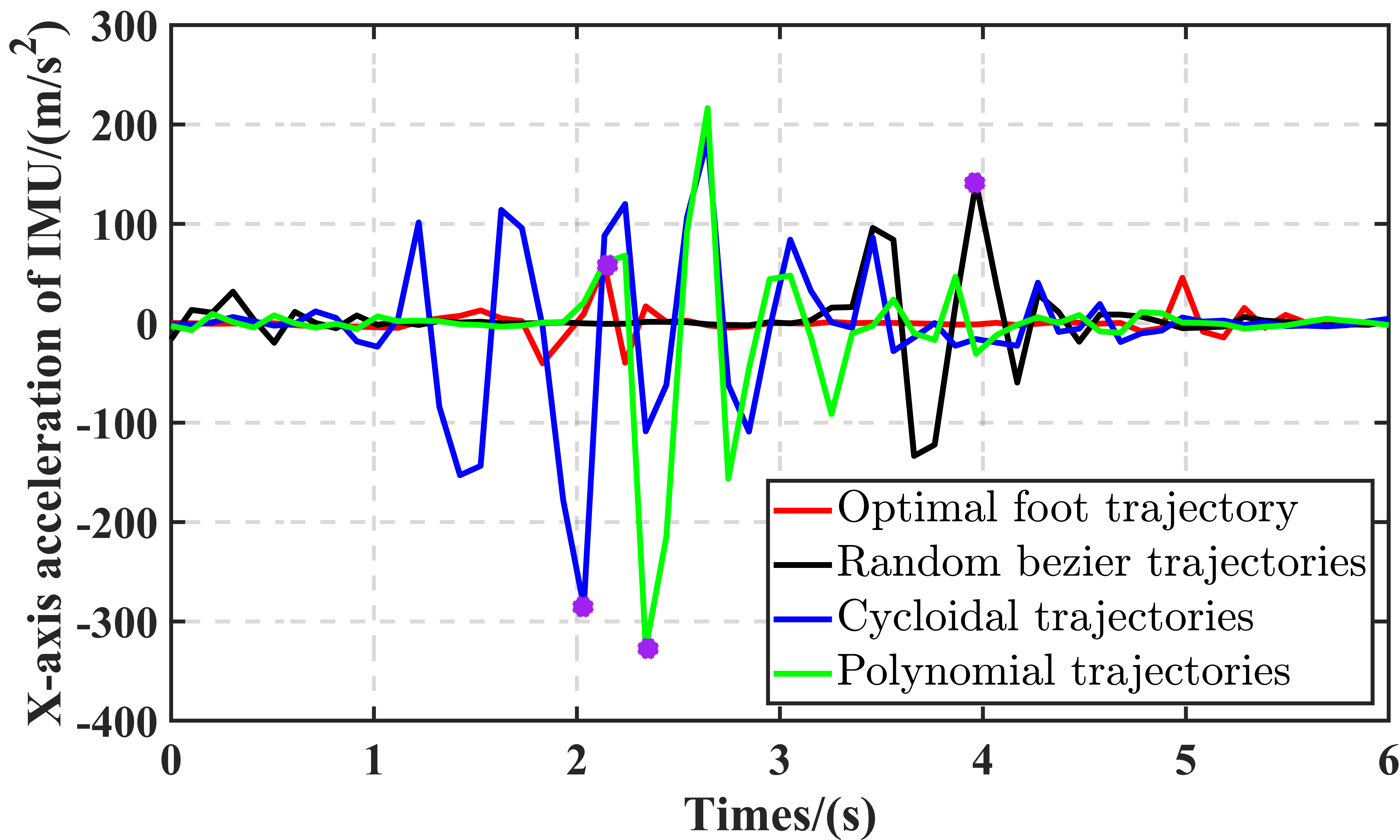}
	\caption{Jitter Data for Optimal Foot Trajectory and Common Foot Trajectory Running on MST-M3F}
	\label{fig:IMU_Vibration}
\end{figure}

Secondly, the evaluation was carried out by the actual operation effect on the legged climbing robot. The IMU sensors carried on the body were utilized to test the performance of various types of end-of-foot trajectories during the climbing process, especially the jitter when the end of the foot was disengaged, as shown in the Fig. \ref{fig:IMU_Vibration_climbing}. The specific body jitter data collected is shown in Fig. \ref{fig:IMU_Vibration}.

In order to further verify the performance of the optimal trajectories, we synthesize and compare the maximum detachment force and maximum pre-pressure data obtained in the single-leg motion test with the maximum jitter data obtained in the fuselage jitter test, and the results are shown in Table \ref{tab:Foot_Trajectory_Performance}.

\begin{table}[!htbp]
	\caption{Comparison of the performance of various types of foot trajectories}
	\label{tab:Foot_Trajectory_Performance}
	\centering
	\footnotesize 
	\setlength{\tabcolsep}{1.5pt}
	\renewcommand{\arraystretch}{1.2}  
	\begin{tabular}{ccccc}
	\toprule
		Contrast index & Polynomial & Cycloidal & Random bezier & Optimal  \\
	\midrule
	Max detachment force(N) & 7.49 & 7.24 & 6.11 & 5.40 \\
	Max pre-pressure(N) & 5.41 & 5.21 & 5.85 & 5.85 \\
	Max body jitter(m/s$^2$) & 327.27 & 285.22 & 141.30 & 58.41 \\
	\bottomrule
	\end{tabular}
\end{table}

The results show that the optimized trajectory obtained by using the FTFOF algorithm is superior to other commonly used foot-end trajectories in all aspects. Especially for the maximum detachment force and body shake data, the maximum detachment force of the optimal foot-end trajectory is reduced by 28 $\%$ compared with the polynomial trajectory; the maximum body shake of the optimal foot-end trajectory is reduced by 82 $\%$ compared with the polynomial trajectory. Therefore, the optimal end-of-foot trajectory obtained by the FTFOF algorithm in this paper is superior in performance and can meet the requirements of stable climbing motion.

\section{FTFOF Algorithm Generalization Verification Experiment}
\label{sec:Experimental Results and Analysis}

In order to verify the generalization index of the FTFOF algorithm proposed in this paper, a quadrupedal climbing robot MST-Q~\cite{xiao_MST-Q}, which is completely different from MST-M3F, is selected for testing. If the algorithm can be quickly ported to MST-Q with good results, it will be a strong proof of the generality of the FTFOF algorithm. For the sake of comparison, the most commonly used straight line climbing foot-end trajectory is still taken as an example for verification. The specific testing procedure is the same as that of MST-M3F, with the only difference that different minimum detachment point constraints and trajectory shape constraints need to be reset for the unique foot-end structure of MST-Q.

\begin{figure}[h]   
	\centering
    \includegraphics[width=0.2\textwidth]{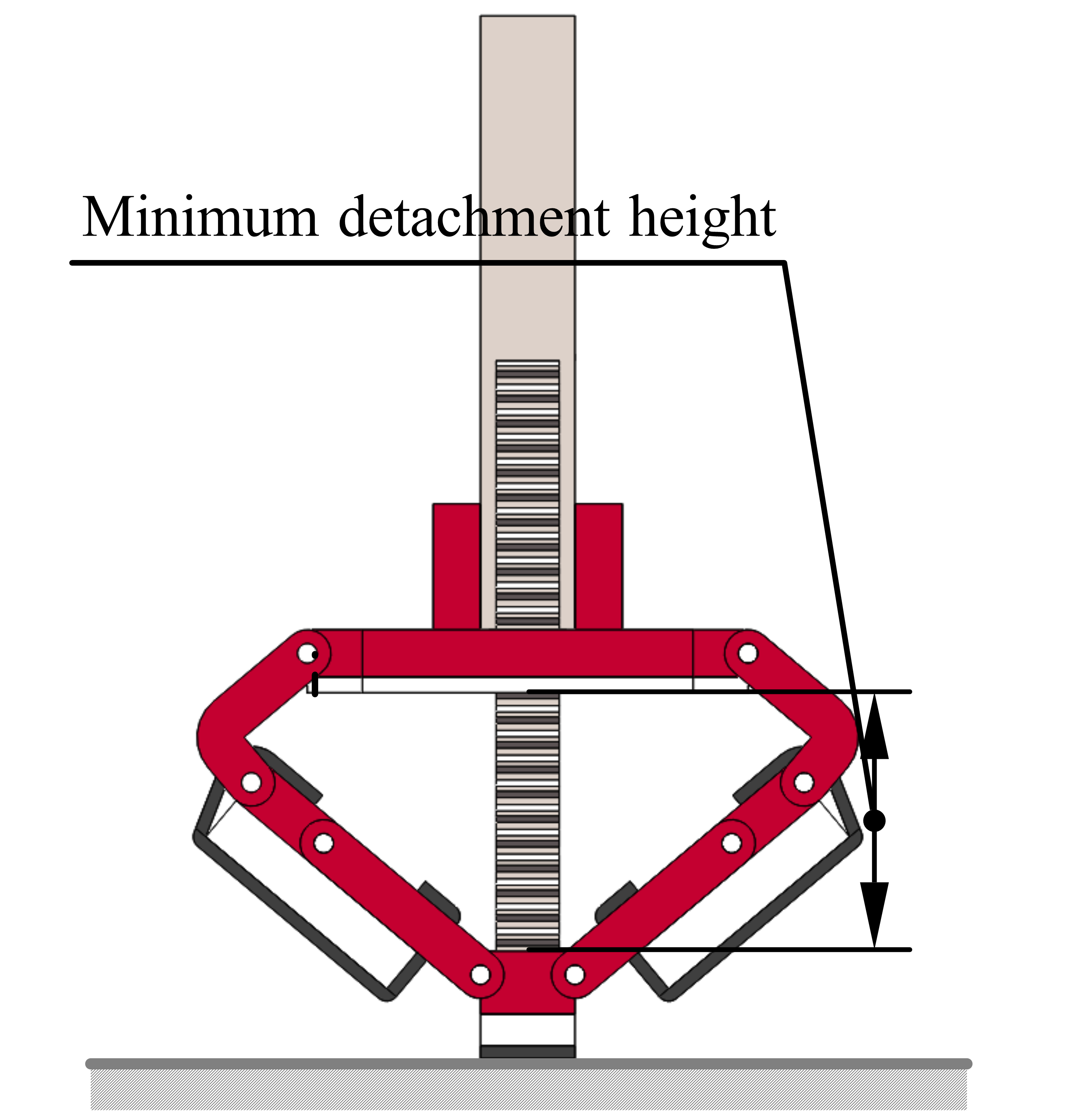}
	\caption{MST-Q Minimum detachment height to achieve detachment motion.}
	\label{fig:MST_Q_min_height}
\end{figure}

Since the MST-Q needs to move vertically to achieve detachment, as shown in Fig. \ref{fig:MST_Q_min_height}. Therefore, we can directly set the minimum detachment point of MST-Q $m_d$ = (0, 0.019) (m), i.e., the minimum detachment height of MST-Q to realize detachment is 0.019 m. 

In order to ensure the stable adherence of MST-Q, the adhering foot will move down a certain distance after it reaches the surface of the adhering surface. Due to the synergistic mechanism of the drive and support racks, this movement does not affect the adhesion effect, but rather enhances the adhesion force. Therefore, it is necessary to impose constraints on control points $P_{10}$, $P_{11}$ and $P_{12}$ to ensure that the adhering foot can continue to move down after reaching the adhering surface without displacement in the horizontal direction. The specific constraints are $P_{10}=P_{11} =P_{12}$. the other constraints are the same as for MST-M3F.

\begin{figure}[h]   
	\centering
    \includegraphics[width=0.48\textwidth]{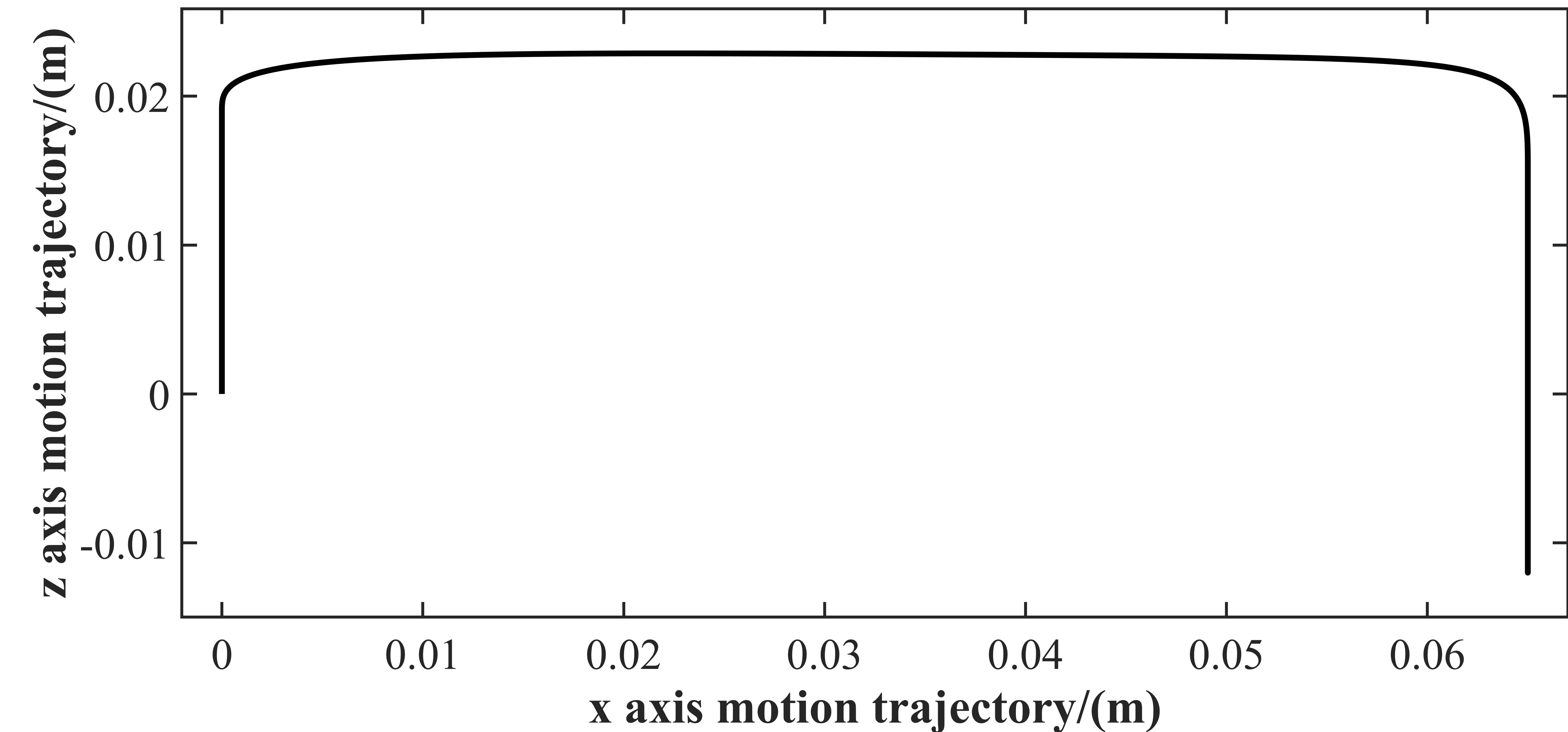}
	\caption{MST-Q foot running curve.}
	\label{fig:foot_traj}
\end{figure}

Considering the unique vertical detachment characteristics of MST-Q, in the optimization process, in addition to the maximum detachment force constraint strategy $f_{s1}$, the adhesion performance constraint strategy $f_{s3}$, the foot-end jitter constraint strategy $f_{s7}$ and the average detachment force constraint strategy $f_{s2}$ as the objective function, in addition to the foot-end lifting height constraint strategy $f_{s4}$. In addition, the trajectory bending constraint strategy $f_{s6}$ remains as the equation constraint and solves for the corresponding Pareto front. The specific objective functions are not repeated. And these objective functions are prioritized as $f_{s1}$, $f_{s3}$, $f_{s7}$,$f_{s3}$, $f_{s2}$, and selects the optimal solution that meets the requirements from the Pareto frontiers $P_f$ through a redundant hierarchical strategy [0, 0, 0, 65.0,65.0, 65.0, 10.0, 15.0, 19.0, 15.0, 1.0, -38.34], and finally obtain the optimal trajectory of the foot end, as shown in Fig. \ref{fig:foot_traj}.

Comparing the obtained optimal foot trajectories with the commonly used foot trajectories, an obvious problem can be found: due to the unique vertical detachment characteristics of the MST-Q robot, the existing commonly used foot trajectories cannot meet the relevant constraints, and thus cannot complete the climbing task. In contrast, the optimal foot-end trajectory proposed in this paper can generate the required optimal foot-end trajectory to complete the climbing motion due to the targeted optimization of foot-end force and foot-end trajectory.

\begin{figure}[h]   
	\centering
    \includegraphics[width=0.48\textwidth]{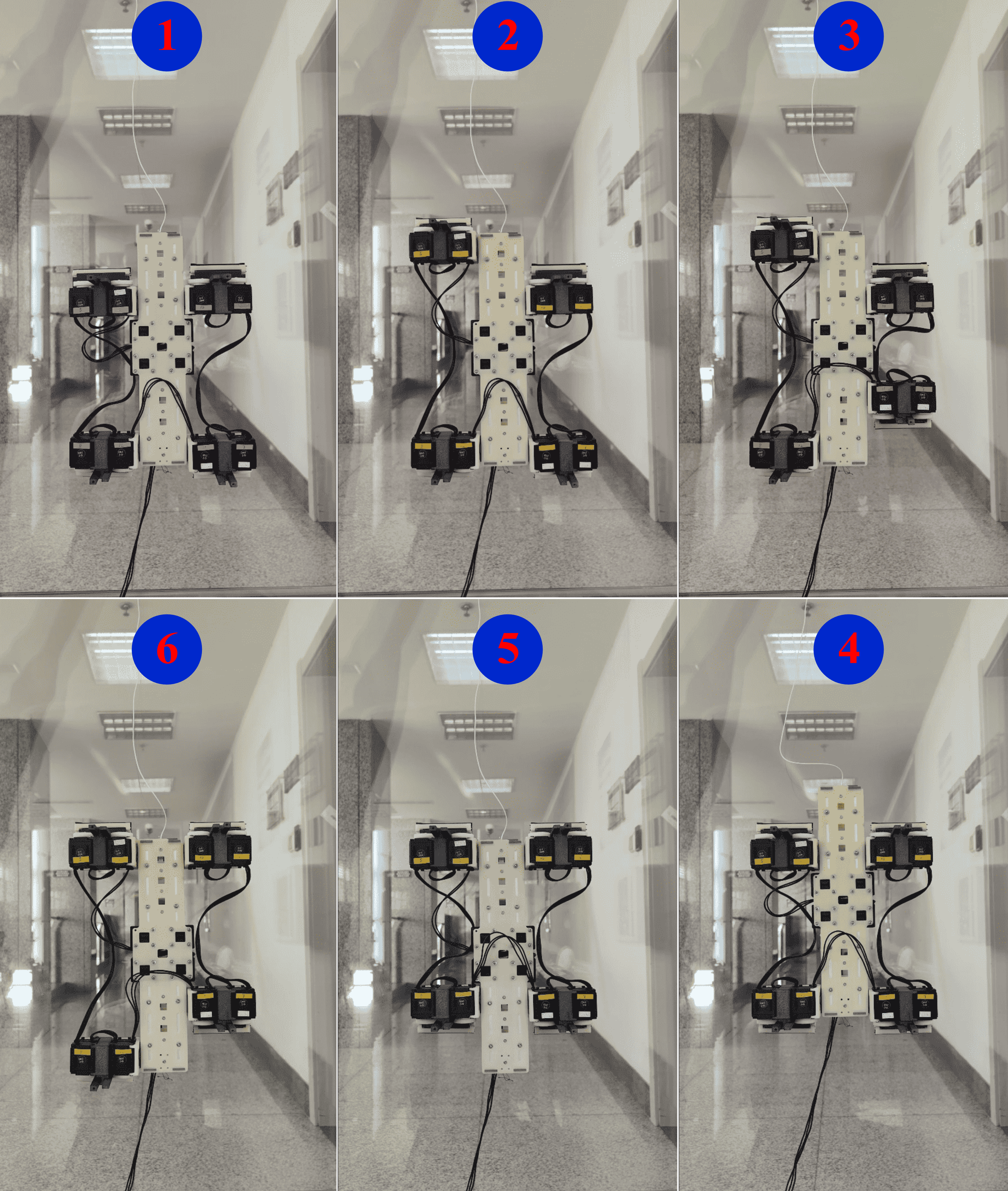}
	\caption{MST-Q performs straight climbing movements with optimal foot trajectories.}
	\label{fig:MST_Q_climb}
\end{figure}

In order to better highlight the versatility of the FTFOF algorithm, the optimal end-of-foot trajectory obtained in this paper is subjected to actual climbing test experiments on the MSTQ. In addition, in order to differentiate from the test environment of MST-M3F, the glass surface is chosen as the climbing environment in this experiment, and the specific experiment is shown in Fig. \ref{fig:MST_Q_climb}.

Based on the results of the above experiments and analyses, it can be seen that the foot trajectories obtained based on the FTFOF algorithm proposed in this paper can be quickly transplanted to other types of dry-adhesive foot climbing robots, and have the versatility of the algorithm.

\section{Conclusion}
\label{sec:conclusion}

In this paper, we propose an optimization algorithm framework for regulating the foot detachment force and adhesion force through the foot trajectory: the Foot Trajectory and Force Optimization Framework for Footed Climbing Robots. Based on the multi-modal training set obtained from the mobile foot force generalized data acquisition platform, the framework is able to generate foot trajectories that meet the expectations according to different task requirements, ensuring that the robot achieves high adhesion, low detachment, and low jitter performances during the movement process, and thus realizes a stable climbing motion. To further improve the efficiency, this chapter proposes a redundant grading strategy, which can quickly find the optimal trajectory in the Pareto front that best meets the current task requirements. Finally, the effectiveness of the framework is verified through the experimental comparison between the optimal foot-end trajectories and the commonly used trajectories. The results show that the maximum detachment force of the optimal foot trajectory is reduced by 28 \% compared with the polynomial trajectory, and the maximum body jitter is reduced by 82 \% compared with the polynomial trajectory. In addition, the generality and portability of the framework is demonstrated by the fast implementation of the algorithm on MST-Q, a quadrupedal climbing robot with completely different foot structures, with good results. Therefore, the optimal foot trajectories obtained by the FTFOF algorithm proposed in this paper are superior in performance and can meet the requirements of stable climbing motion.

\bibliographystyle{IEEEtran}
\bibliography{references}

\begin{thebibliography}{10}
\providecommand{\url}[1]{#1}
\csname url@rmstyle\endcsname
\providecommand{\newblock}{\relax}
\providecommand{\bibinfo}[2]{#2}
\providecommand\BIBentrySTDinterwordspacing{\spaceskip=0pt\relax}
\providecommand\BIBentryALTinterwordstretchfactor{4}
\providecommand\BIBentryALTinterwordspacing{\spaceskip=\fontdimen2\font plus
\BIBentryALTinterwordstretchfactor\fontdimen3\font minus \fontdimen4\font\relax}
\providecommand\BIBforeignlanguage[2]{{%
\expandafter\ifx\csname l@#1\endcsname\relax
\typeout{** WARNING: IEEEtran.bst: No hyphenation pattern has been}%
\typeout{** loaded for the language `#1'. Using the pattern for}%
\typeout{** the default language instead.}%
\else
\language=\csname l@#1\endcsname
\fi
#2}}

\bibitem{li2022design}
Z.~Li, Z.~Li, L.~M. Tam, and Q.~Xu, ``Design and development of a versatile quadruped climbing robot with obstacle-overcoming and manipulation capabilities,'' \emph{IEEE/ASME Transactions on Mechatronics}, vol.~28, no.~3, pp. 1649--1661, 2022.

\bibitem{murphy2011waalbot}
M.~P. Murphy, C.~Kute, Y.~Meng{\"u}{\c{c}}, and M.~Sitti, ``Waalbot ii: Adhesion recovery and improved performance of a climbing robot using fibrillar adhesives,'' \emph{The International Journal of Robotics Research}, vol.~30, no.~1, pp. 118--133, 2011.

\bibitem{gitardi2023trajectory}
D.~Gitardi, S.~Sabbadini, and A.~Valente, ``Trajectory error compensation for optimal control of uma-2--a climbing robot executing maintenance operation in harsh environment,'' in \emph{2023 IEEE International Conference on Robotics and Automation (ICRA)}.\hskip 1em plus 0.5em minus 0.4em\relax IEEE, 2023, pp. 10\,090--10\,096.

\bibitem{shi2022active}
Y.~Shi, Z.~Gong, B.~Tao, Z.~Yin, and H.~Ding, ``An active compliance adsorption method for climbing machining robot on variable curvature surface,'' \emph{IEEE/ASME Transactions on Mechatronics}, vol.~28, no.~2, pp. 1127--1136, 2022.

\bibitem{tang2018switchable}
Y.~Tang, Q.~Zhang, G.~Lin, and J.~Yin, ``Switchable adhesion actuator for amphibious climbing soft robot,'' \emph{Soft robotics}, vol.~5, no.~5, pp. 592--600, 2018.

\bibitem{zi2024intelligent}
P.~Zi, K.~Xu, J.~Chen, C.~Wang, T.~Zhang, Y.~Luo, Y.~Tian, L.~Wen, and X.~Ding, ``Intelligent rock-climbing robot capable of multimodal locomotion and hybrid bioinspired attachment,'' \emph{Advanced Science}, p. 2309058, 2024.

\bibitem{nadan2024loris}
P.~Nadan, S.~Backus, and A.~M. Johnson, ``Loris: A lightweight free-climbing robot for extreme terrain exploration,'' in \emph{2024 IEEE International Conference on Robotics and Automation (ICRA)}.\hskip 1em plus 0.5em minus 0.4em\relax IEEE, 2024, pp. 18\,480--18\,486.

\bibitem{bandyopadhyay2018magneto}
T.~Bandyopadhyay, R.~Steindl, F.~Talbot, N.~Kottege, R.~Dungavell, B.~Wood, J.~Barker, K.~Hoehn, and A.~Elfes, ``Magneto: A versatile multi-limbed inspection robot,'' in \emph{2018 IEEE/RSJ International Conference on Intelligent Robots and Systems (IROS)}.\hskip 1em plus 0.5em minus 0.4em\relax IEEE, 2018, pp. 2253--2260.

\bibitem{hong2022agile}
S.~Hong, Y.~Um, J.~Park, and H.-W. Park, ``Agile and versatile climbing on ferromagnetic surfaces with a quadrupedal robot,'' \emph{Science Robotics}, vol.~7, no.~73, p. eadd1017, 2022.

\bibitem{sriratanasak2022tasering}
N.~Sriratanasak, D.~Axinte, X.~Dong, A.~Mohammad, M.~Russo, and L.~Raimondi, ``Tasering twin soft robot: A multimodal soft robot capable of passive flight and wall climbing,'' \emph{Advanced Intelligent Systems}, vol.~4, no.~12, p. 2200223, 2022.

\bibitem{gu2018soft}
G.~Gu, J.~Zou, R.~Zhao, X.~Zhao, and X.~Zhu, ``Soft wall-climbing robots,'' \emph{Science Robotics}, vol.~3, no.~25, p. eaat2874, 2018.

\bibitem{hernando2022romerin}
M.~Hernando, E.~Gambao, C.~Prados, D.~Brito, and A.~Brunete, ``Romerin: A new concept of a modular autonomous climbing robot,'' \emph{International Journal of Advanced Robotic Systems}, vol.~19, no.~5, p. 17298806221123416, 2022.

\bibitem{kalouche2014inchworm}
S.~Kalouche, N.~Wiltsie, H.-J. Su, and A.~Parness, ``Inchworm style gecko adhesive climbing robot,'' in \emph{2014 IEEE/RSJ International Conference on Intelligent Robots and Systems}.\hskip 1em plus 0.5em minus 0.4em\relax IEEE, 2014, pp. 2319--2324.

\bibitem{li2022robust}
X.~Li, P.~Bai, X.~Li, L.~Li, Y.~Li, H.~Lu, L.~Ma, Y.~Meng, and Y.~Tian, ``Robust scalable reversible strong adhesion by gecko-inspired composite design,'' \emph{Friction}, vol.~10, no.~8, pp. 1192--1207, 2022.

\bibitem{GRU}
K.~{Cho}, ``Learning phrase representations using rnn encoder-decoder for statistical machine translation,'' \emph{arXiv preprint arXiv:1406.1078}, 2014.

\bibitem{DILATE}
V.~{Le Guen} and N.~{Thome}, ``Shape and time distortion loss for training deep time series forecasting models,'' \emph{Advances in neural information processing systems}, vol.~32, 2019.

\bibitem{NSGAII}
K.~{Deb}, A.~{Pratap}, S.~{Agarwal}, and T.~{Meyarivan}, ``A fast and elitist multiobjective genetic algorithm: Nsga-ii,'' \emph{IEEE transactions on evolutionary computation}, vol.~6, no.~2, pp. 182--197, 2002.

\bibitem{ARIMA}
G.~E. {Box}, G.~M. {Jenkins}, G.~C. {Reinsel}, and G.~M. {Ljung}, \emph{Time series analysis: forecasting and control}.\hskip 1em plus 0.5em minus 0.4em\relax John Wiley \& Sons, 2015.

\bibitem{Exponential_Smoothing}
D.~J. {Hand}, ``Forecasting with exponential smoothing: the state space approach by rob j. hyndman, anne b. koehler, j. keith ord, ralph d. snyder,'' 2009.

\bibitem{pymoo}
J.~Blank and K.~Deb, ``Pymoo: Multi-objective optimization in python,'' \emph{Ieee access}, vol.~8, pp. 89\,497--89\,509, 2020.

\bibitem{xiao_MST-Q}
J.~{Xiao}, L.~{Hao}, H.~{Xu}, X.~{Zhang}, X.~{Li}, and Z.~{Li}, ``Mst-q: Micro suction tape quadruped robot with high payload capacity,'' \emph{IEEE Robotics and Automation Letters}, vol.~8, no.~11, pp. 7304--7311, 2023.

\end{thebibliography}

\end{document}